\documentclass[]{article}

\usepackage{microtype}
\usepackage[pdftex]{graphicx}
\usepackage{float}
\usepackage{subfigure}
\usepackage{booktabs} 

\usepackage{hyperref}

\usepackage[accepted]{icml2022}

\usepackage{amsmath}
\usepackage{amssymb}
\usepackage{mathtools}
\usepackage{amsthm}

\usepackage[capitalize,noabbrev]{cleveref}

\theoremstyle{plain}

\theoremstyle{definition}

\theoremstyle{remark}

\usepackage[textsize=tiny]{todonotes}

\newcommand{\TLP}{\emph{TLP} }

\newcommand{\rtest}{R$_{\mbox{test}}$}
\newcommand{\GPT}{\mbox{GPT-2}}

\icmltitlerunning{Neural Language Models are not Born Equal to Fit Brain Data, but Training Helps}

\begin{document}

\twocolumn[

\icmltitle{Neural Language Models are not Born Equal to Fit Brain Data, but Training Helps}


\icmlsetsymbol{equal}{*}

\begin{icmlauthorlist}
\icmlauthor{Alexandre Pasquiou}{equal,unicog,parietal}
\icmlauthor{Yair Lakretz}{equal,unicog}
\icmlauthor{John Hale}{uga}
\icmlauthor{Bertrand Thirion}{parietal}
\icmlauthor{Christophe Pallier}{unicog}
\end{icmlauthorlist}

\icmlaffiliation{unicog}{Cognitive Neuroimaging Unit, INSERM, CEA, Neurospin, Gif-sur-Yvette, France}
\icmlaffiliation{parietal}{Parietal, INRIA, CEA, Neurospin, Gif-sur-Yvette, France}
\icmlaffiliation{uga}{Dept. of Linguistics, U. of Georgia, Athens, GA, USA}

\icmlcorrespondingauthor{Alexandre Pasquiou}{alexandre.pasquiou@inria.fr}

\icmlkeywords{Machine Learning, ICML, fMRI, NLP, LSTM, BERT, GPT-2, Perplexity, neuroimaging}

\vskip 0.3in
]



\printAffiliationsAndNotice{\icmlEqualContribution} 

\begin{abstract}
Neural Language Models (NLMs) have made tremendous advances during the last years, achieving impressive performance on various linguistic tasks.
Capitalizing on this, studies in neuroscience have started to use NLMs to study neural activity in the human brain during language processing.
However, many questions remain unanswered regarding which factors determine the ability of a neural language model to capture brain activity (aka its 'brain score').
Here, we make first steps in this direction and examine the impact of test loss, training corpus and model architecture (comparing GloVe, LSTM, \GPT{} and BERT), on the prediction of functional Magnetic Resonance Imaging timecourses of participants listening to an audiobook.
We find that (1) untrained versions of each model already explain significant amount of signal in the brain by capturing similarity in brain responses across identical words, with the untrained LSTM outperforming the transformer-based models, being less impacted by the effect of context; (2) that training NLP models improves brain scores in the same brain regions irrespective of the model's architecture; (3) that Perplexity (test loss) is not a good predictor of brain score; (4) that training data have a strong influence on the outcome and, notably, that off-the-shelf models may lack statistical power to detect brain activations. 
Overall, we outline the impact of model-training choices, and suggest good practices for future studies aiming at explaining the human language system using neural language models.

\end{abstract}

\section{Introduction}\label{introduction}

In the last few years, Transformer-based language models have revolutionized the field of natural language processing in virtually all areas. Although these models were developed for applications in language technology, their impressive success has raised interest in whether these models could also shed light on language processing in the human brain. Promising results in this direction suggest that brain activations and transformer-based models converge to similar linguistic representations \cite{caucheteux:icml2021} showing that brain activity can be significantly predicted from linear combinations of model activations, as was shown for fMRI \citep{toneva:neurips2020,caucheteux:icml2021,caucheteux:emnlp2021}, MEG \citep{Caucheteux:2021c}, and intracranial data \cite{Goldstein:2020}.

However, several differences between Transformer-based models and the human brain raise questions about how far we can advance our understanding of brain function using these models. First, the architecture of Transformers is based on multi-head self-attention modules, which does not clearly map on neural computations in biological networks \citep[e.g.,][]{dayan2005theoretical}. Does this architecture contribute to or hinder the ability of the model to predict brain activity compared to other, possibly more brain-like, architectures (e.g., recurrent neural networks)? Second, the data used to train Transformer-based models is often different from that available for children, both in type and size. Training a Transformer-based model requires massive corpora, on the order of billions of words, whereas children require orders of magnitudes less words to achieve comparable or better linguistic performance. How does the training corpus (type and size) affect the model's ability to fit brain activity? Finally, the learning and evaluation objective commonly used with these models, such as masked or next-word prediction, is at most a rough approximation of the computational problem the human brain solves during language acquisition and processing. Can one consider that a well-trained model (according to perplexity loss) is a good model for brain activity in language tasks?

We investigate these questions by contrasting several types of language models in their ability to fit functional Magnetic Resonance Imaging (fMRI) timecourses of participants listening to the `The Little Prince’ audiobook. 
Importantly, we conduct the model comparison while controlling for various aspects of the architecture of the models and the type and size of the corpus on which they are trained. 
To address the first question about the architecture of the models, we study the ability of untrained models to fit brain activity. 
We obtain significant differences across architectures, with that of recurrent neural networks achieving highest scores. 
Next, we study brain-score gains brought by training across models, and find a network of brain regions, in which brain activity is consistently better fitted by various types of models. 
Moreover, running a comprehensive comparison of neural language models, we find that the effect of training is stronger in the case of Transformer-based models. 
We next question the relationship between perplexity and brain score, and study it across models and across training epochs during convergence. In contrast to previous studies, we find that perplexity is not a reliable predictor of model's brain score. 
%
Finally, we show the impact of training data on the model ability to fit brain data,  notably, that off-the-shelf models, such as ones trained only on Wikipedia, may lack statistical power to fit brain activation.

Taken together, we conclude that while the starting point of Transformer-based models is less advantageous compared to that of recurrent neural networks, due to differences in their architectures, training leads to them outperforming all other models in predicting brain data.

\section{Related Literature}
Current knowledge about the cerebral basis of language mostly comes from brain imaging studies that have used tightly controlled stimuli, typically isolated words or sentences out of context \citep[see][for reviews]{price_review_2012,hickok_neurobiology_2015}. As conclusions from such studies may be bounded to the peculiarity of the task and setup used in the experiment \citep{varoquaux2019predictive}, researchers have become more and more interested in data using ``Ecological Paradigms'', in which participants are engaged in more natural tasks, such as conversation or story listening \citep[e.g.][]{regev_selective_2013,lerner_topographic_2011,wehbe_simultaneously_2014}.

Ecological paradigms commonly require methodologies of machine learning based on predictive modeling, to account for the high number of degrees of freedom in the complex system that is the brain. It has been shown that representations extracted from computational models can explain part of the signal acquired in brain neuroimaging. This was shown in early studies by using non-contextualized semantic representations \citep{mitchell_predicting_2008, huth_natural_2016}, moving in later studies to recurrent neural networks to extract context-based word representations \citep{jain_incorporating_2018, jain2021interpretable}, and more recently to Transformer-based language models \citep[e.g.,][]{toneva:neurips2020, caucheteux:icml2021, caucheteux:emnlp2021, Goldstein:2020} - see \citet{hale2022neurocomputational} for a review. 

Interestingly, the architecture of neural language models has been shown to substantially contribute to the ability of the model to fit brain data. Untrained neural language models fitted human brain activity surprisingly well \citep{schrimpf_artificial_2020}. Training was shown to improve brain scores by around 50\% on average, across different architectures. This was suggested as evidence that the human cortex might already provide a sufficiently rich structure for relatively rapid language acquisition. However, conclusions in \citet{schrimpf_artificial_2020} were based on relatively small datasets, from no more than 9 participants. Also, different models in the comparison had different number of units, layers, and were trained on different datasets with varying vocabulary sizes. In our work, we suggest a more controlled study of the effect of architecture and training on brain score, comparing different types of models, while controlling for the aforementioned factors, using a larger brain-imaging dataset, from 51 participants. 

The performance of neural language models on a next-word prediction task, but not on other linguistic tasks, was shown to correlate with their ability to fit human brain data \citep{schrimpf_artificial_2020}. This was suggested as evidence that predictive processing shapes language-comprehension mechanisms in the brain. Here, we question this conclusion and study the relation between next-word prediction and brain score in various types of models, training corpora and training steps.

\section{Analysis Setting: Fitting Brain Data with Modern NLP Models}
\label{preliminaries}

Investigating the ability of neural language model to capture brain activity, we (1) first defined the three families of model architectures that we explored: non-contextualized word embeddings (GloVe; \citealt{pennington_glove_2014}), a recurrent neural network (LSTMs; \citealt{hochreiter_long_1997}) and two Transformer-based models (\GPT{}; \citealt{radford_language_2019} and BERT; \citealt{devlin_bert_2019}); (2) we then trained and tested each model as described in the following paragraphs; before (3) presenting the story \emph{The Little Prince} \cite{lpp} to both human participants and artificial neural language models. Their activation in response to each word and punctuation sign of \emph{The Little Prince} was extracted and (4) used to fit participants fMRI brain activations thanks to regularized linear encoding models. (5) Finally, at the end of the analysis pipeline, we had for each model: a test loss evaluated on our test set and a volumic R maps containing, for each brain voxel, the cross-validated correlation between the encoding model prediction and the observed fMRI response.

\subsection{Datasets}

\textbf{Brain-imaging data.} The brain data consisted of functional Magnetic Resonance Imaging (fMRI) scans from 51 participants who listened to the entire audiobook of \emph{The Little Prince}\footnote{Available from \url{https://openneuro.org/datasets/ds003643/versions/1.0.2}}.
For each participant, there were 9 runs of fMRI acquisition lasting about 10 minutes.  Whole brain volumes were acquired every 2 seconds. A global brain mask was computed to only keep voxels containing useful signal across all runs for at least 50\% of all participants (26,164 voxels). 
Finally, linear detrending and standardization (mean removal and scaling to unit variance) were applied to each voxel's time-series. 
The analysis pipeline relies on Nilearn (v.0.8.1) for data access and visualization. Encoding and subsequent statistical analyses were run with custom Python code using sklearn.

The acoustic onsets and offsets of the 15,435 spoken words were marked to align the audio recording with the text of \emph{The Little Prince}.
In addition to the words, the token streams fed to the neural language models included punctuation signs (commas, dots, ...).

\textbf{Text Corpora.} We designed several datasets on which we trained and evaluated our models. In total, we created 6 training datasets from Wikipedia (425M) 
and Project Gutenberg \footnote{Project Gutenberg. (n.d.). Retrieved February 21, 2016, from www.gutenberg.org.}, using up to 2 thirds of the entire Project Gutenberg in the \textit{xlarge} version and splitting the remaining 1/3 left into equal size validation (1.1G) and test sets (1.1G). 
The datasets created from the Gutenberg Project's data are nested ($small (240M) \subset medium (737M) \subset large (2.2G) \subset xlarge (4.4G) \subset full (4.8G; xlarge+Wikipedia)$).

\subsection{Pipeline}

\textbf{Models.} Given common training, validation and test datasets, we trained several instances of GloVe, LSTM, \GPT{} and BERT.
\begin{itemize}
\item GloVe was trained using the open-source code made available by Pennington and al. \footnote{https://nlp.stanford.edu/projects/glove/.},
\item GPT-2 and LSTM were trained on a Language Modeling task,
\item while BERT was trained on a Masked-Language Modeling task with a 15\% masking-rate.
\end{itemize}
Each model had a vocabulary of 50,001 tokens.
GloVe and LSTM were trained for 23 epochs, while \GPT{} and BERT were trained during 5 epochs. 
Convergence assessment and comparisons with off-the-shelf models are provided in Supplementary Material.
For computational cost reasons, we limited our analysis to 1, 2 and 4-layers models. In the following, we denoted MODEL.X a MODEL with X-layers.

\begin{figure}[t]\label{Fig:pipeline}
\centering
\includegraphics[width=1\columnwidth]{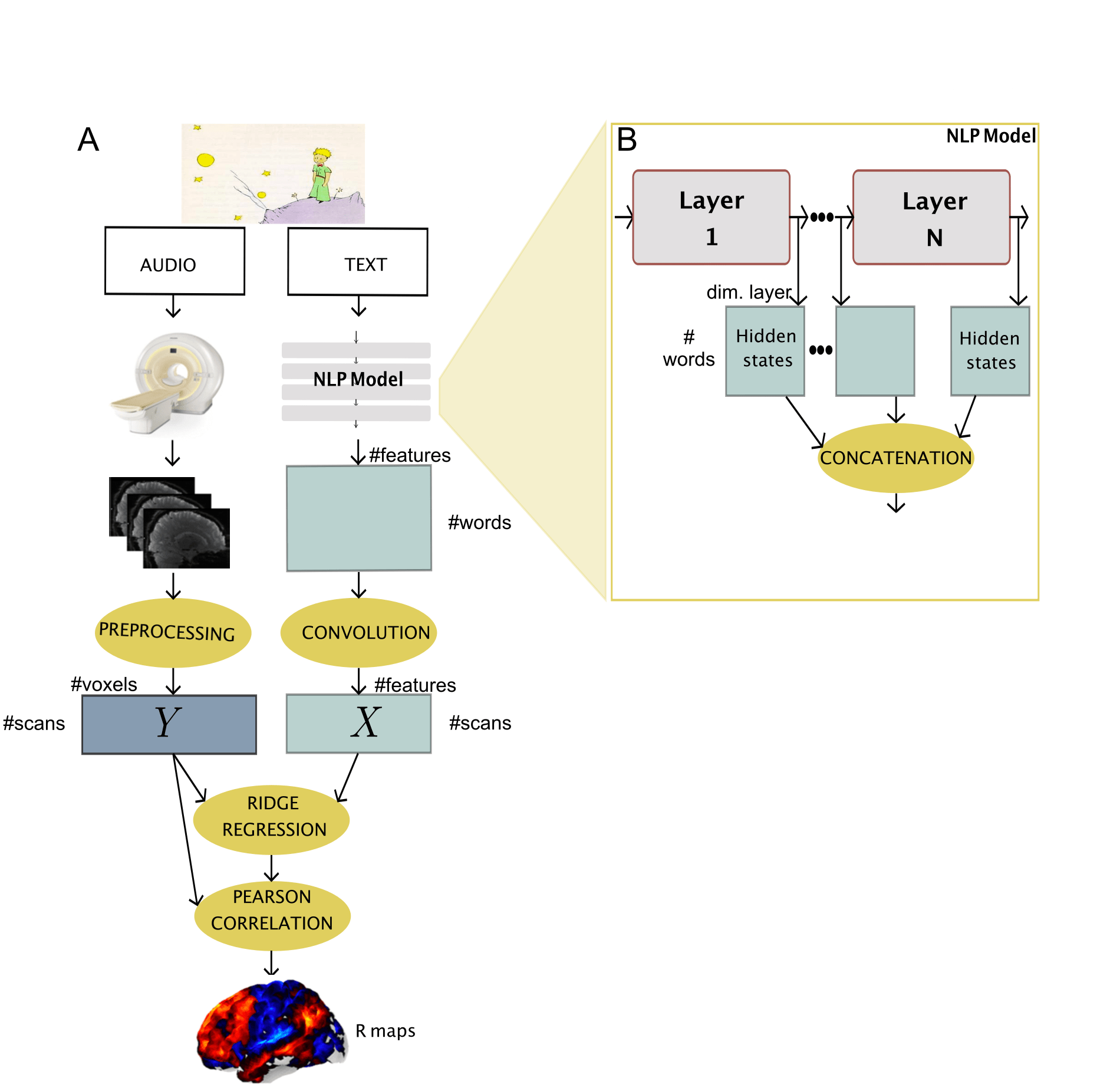}
\caption{\textbf{Overview of the pipeline:} (A) Human participants were presented with an auditory version of \emph{The Little Prince} story while their brain activity was recorded with fMRI. Neural models were presented with a text transcription and the entire state of the network was recorded for each word and punctuation sign. Pre-processing steps were applied to both brain and model activations before aligning the two signals using a nested cross-validated Ridge-regression model. Finally, brain maps of correlation coefficients between models' predictions and fMRI time-series were computed. 
(B) In the case of models with several layers, model activations were extracted from each layer and were concatenated into a single activation matrix.}
\end{figure}

\textbf{Activation generation.} (See Fig.\ref{Fig:pipeline}) We presented the transcription of the audio book used to acquired fMRI brain data (The Little Prince, TLP; 15.435 words) to both trained and untrained versions of the selected artificial neural language models. 

For each model, we extracted the model hidden-states while it processed its input, and defined from it what we call the activation matrix (one for each run). 
More precisely, if we note $d$ the dimensionality of a neural model, which corresponds to the total number of units in the model, and $w$ the total number of words in the text. 
We obtained an activation matrix $A \in \mathbb{R}^{w \times d}$  after the presentation of the entire text to the model. 
This means that each word of \TLP is represented by a $d$-dimensional vector.
Then, model activations were transformed into time-series matched to the fMRI acquisition times, using the following procedure:

\begin{enumerate}
    \item \textbf{Normalization:} To match the scale of activations across layers (for multi-layers models), the activations of each layer were normalized by dividing them by the average L2-norm over words.
    \item \textbf{Convolution:} each column of the resulting activation matrix, mapped onto words' offsets times, was convolved with SPM's canonical Haemodynamic Response Function(HRF; \citealt{friston_statistical_2007}) sampled at 0.2s. The outcome was resampled at 2s to match the repetition time of fMRI acquisition and then mean-centered.
\end{enumerate}

\textbf{Fitting brain data.} The latter stage resulted for each model and each run into a design-matrix of size $\#\mbox{scans} \times d$. Given a neural language model, we gave the associated nine design-matrices to a nested cross-validated L2-regularized univariate linear encoding model to fit the fMRI brain data (of size $\#\mbox{scans} \times \#voxels$). 

The \emph{encoding model} is a function that maps a vector of stimulus features onto brain responses activity \cite{naselaris_encoding_2011}. We denote by $x_t$ the vector of features at time $t$, such as predicted time-courses derived from a language model, and by $y_t^v$ the corresponding brain responses measured at voxel $v$. We learnt a linear voxel-level encoding model using Ridge regression,
whose general solution is given by: 
\begin{equation*}
\hat{\beta}_{\mbox{\small Ridge}}^v = arg\min_{\beta}\sum^{n}_{t=1}(y^v_t-\beta^Tx_t)^2 + \lambda\|\beta\|_{2}^{2}
\end{equation*}
To evaluate model performance and the optimal regularization parameter $\lambda^*$, we used a nested cross-validation procedure: we split each participant's dataset into training, validation and test sets, such that the training set included 7 out of the 9 experiment runs, and the validation and test sets contained one of the two remaining sessions. We evaluated model performance using the Pearson correlation coefficient $R$, which is a measure of the linear correlation between models' predicted time-courses and the actual time-courses. It is defined as:
\begin{equation*}
R(y, \hat{y})_{v, test} = \frac{\sum (\hat{y}_t^v - \bar{\hat{y}}^v) (y^v_t - \bar{y}^v)}
         {\sqrt{\sum (\hat{y}_t^v - \bar{\hat{y}}^v)^2 \sum (y^v_t - \bar{y}^v)^2}},
\end{equation*}
\begin{equation*}
\text{where }
\bar{\hat{y}}^v = \frac{1}{T} \sum_{t=1}^{T} \hat{y}_t^v 
\quad \textrm{,} \quad 
\bar{y}^v = \frac{1}{T} \sum_{t=1}^{T} y^v_t
\end{equation*}
For each subject and each voxel, we first determined  $\lambda^*$ by comparing $R_{valid}$ for 10 different values of $\lambda$, linearly spaced in log-scale between $10$ and $10^5$. We then calculated $R_{test}$ for  $\lambda^*$. Finally, we repeated this procedure 9 times, using cross-validation. This resulted in 9 $R_{test}$ values that we then averaged to produce a single $R_{test}$ map for the participant.

\textbf{Results of the analysis pipeline.} Finally, at the end of the analysis pipeline we had for each model: a test loss evaluated on our test set derived from the Gutenberg Project, and a volume-based R map displaying for each brain voxel the correlation between the encoding model prediction and the observed time series. Volume maps are rendered on cortical surfaces by projection.

\section{Methods and Experimental Setup}

\subsection{Assessing model fitness to brain data}
\label{method:reporting_fmri_results}

Whole-brain, voxel-based,  group analyses were performed, using one-sample t-tests applied to the individuals' $R_{test}$ maps spatially smoothed with an isotropic Gaussian kernel with 6mm FWHM. 
In each voxel, the test assessed whether the distribution of $R_{test}$ values across participants was significantly larger than zero. 
To control for multiple comparisons, all the maps displayed in this document were corrected using a Bonferroni correction of 0.1 \cite{bonferroni}, that is, values reported on the maps (e.g. R scores) are shown only for voxels that survived this threshold.

\begin{figure*}[htb]
    \centering
    \includegraphics[width=.8\textwidth]{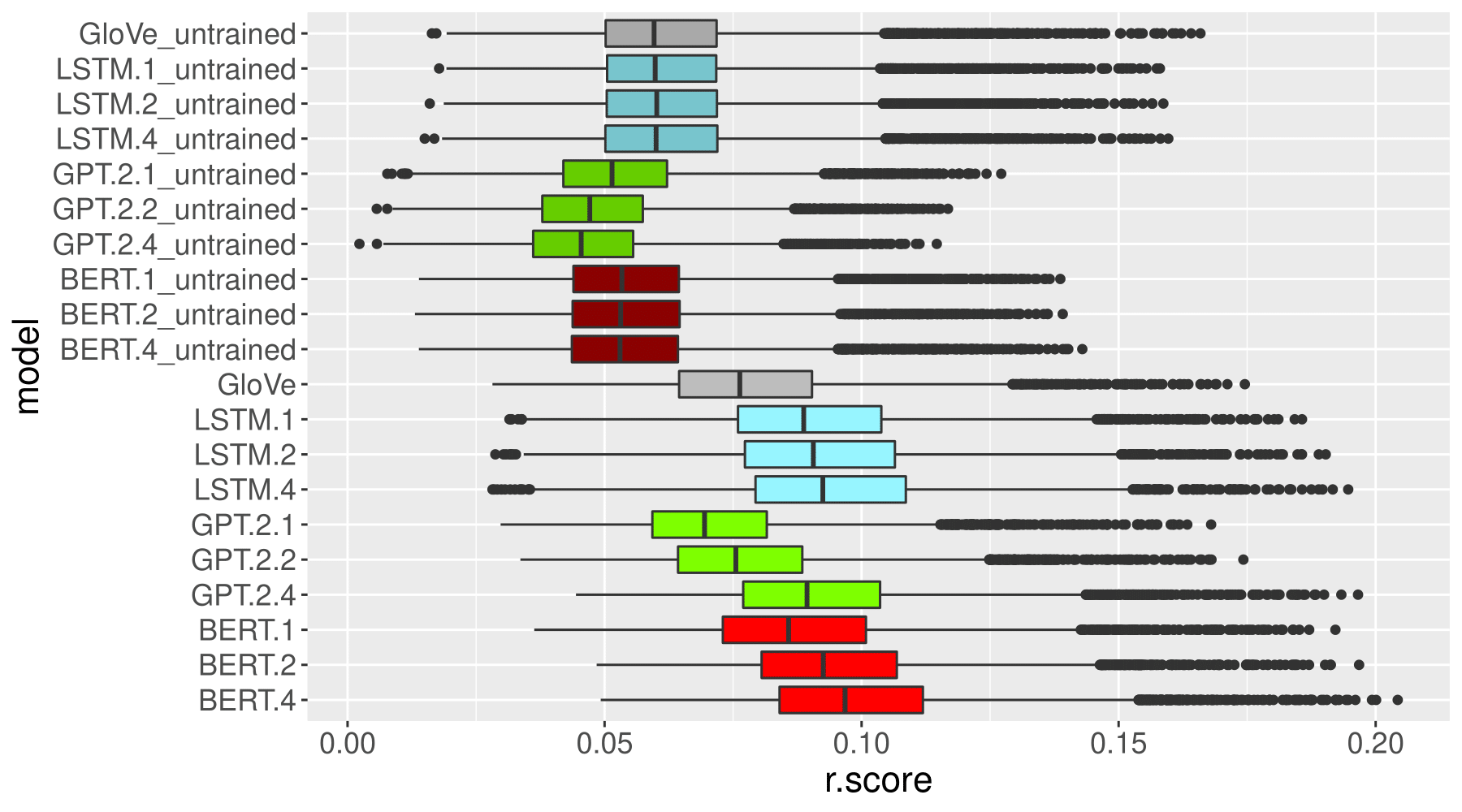}
    \vskip -.05in
    \includegraphics[width=.5\textwidth]{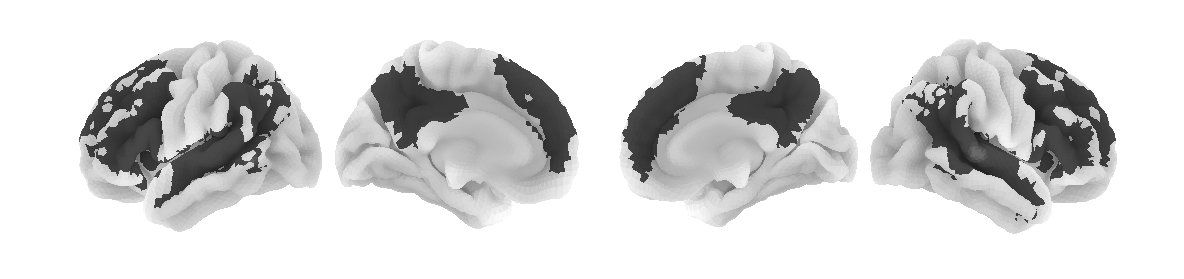}
    \caption{\textbf{Distributions of R$_{\mbox{test}}$-values across voxels in the 25\% most reliable voxels across subjects (mask computed from shared response model, shown below the graphic)} for untrained and trained versions of GloVe (static word embedding), LSTM, \GPT{} and BERT models, having 1, 2 or 4 layers. }
    \label{fig:rvalue-distribs}
\end{figure*}

We derived in a model-agnostic manner from a Shared-Response Model (SRM, \citealt{Chen2015}) the most “responsive” voxels, that is, the voxels whose R values were the 25\% highest ones. This set of 6541 voxels, which we will refer to as “SRM25” is displayed on a brain surface at the bottom of Fig.\ref{fig:rvalue-distribs}. It is used to compute the distributions of brain scores.

\subsection{Comparison of untrained models and baselines}
\label{method:comparing_models}
In our first analysis, we assessed whether the model class and number of layers bias its ability to fit fMRI brain data.
We instantiated several untrained versions of each model class, varying the number of layers, and generated activation matrices from these models before fitting them to brain data. For each model, the activation matrices were built using all the hidden-states of all layers, including the embedding layer.
We also defined a Baseline model whose activation matrices are obtained by associating a fixed embedding vector of size 768 (size of each model's layer) to each word of the text. It is equivalent to an untrained GloVe model (and will be referred to as such).
For each, we obtained 3D brain maps displaying the average $R_{test}$ values in each voxel.
Finally we displayed boxplots of the $R_{test}$ values distributions in the previously SRM-defined voxel selection.

\subsection{Comparison of trained and untrained neural language models}
\label{method:comparing_trained_and_untrained}

We then investigated how training models impacts their ability to fit fMRI brain data.
We first generated activation matrices from the trained language models before fitting them to brain data and finally displaying the group-level difference between each model's map and its untrained version's map.
To study the overlap between the different contrast maps, we extracted for LSTM, \GPT{} and BERT the set of voxels whose R values were the 10\% highest. Then we computed the intersection of these three sets and the percentage of overlap for pairs of models and for the three models.
We additionally identified the set of voxels whose R values were the 10\% highest in the untrained models maps, and computed the intersection of the three sets and the percentage of overlap for the three models.
Finally, we studied the intersection between the 2 overlaps of the three models and synthesized differences and similarities in Fig.\ref{Fig:training_effect}.

\subsection{Mapping the brain scores of NLP models}
\label{method:mapping}

The next step was to run a comprehensive comparison of the selected models to understand models similarities and specificities.
We contrasted individual maps between pairs of models to test in each voxel:
(1) the effect of incorporating context into target-word representations, by contrasting LSTM and GloVe;
(2) the effect of attention vs. recurrence mechanisms on prediction, by contrasting the maps of transformers and LSTM.
(3) the interaction between model architecture and training between transformers and LSTM.
%
(4) and finally the effect of bi-directionality vs. incrementality (BERT vs. \GPT{}). 
Compared models always have the same number of hidden-states.

\subsection{Perplexity and Brain score}
\label{method:perplexity_and_brain_score}

Finally, we investigated the relation between \emph{perplexity} and \emph{brain score}. Using the set of trained LSTM, \GPT{} and BERT models, we evaluated them using the standard loss, that is, the average logarithm of model perplexity, on the \textit{test set}. For each model, we also computed the \emph{brain score}, defined as the average R-value within the SRM25 voxelset. 

We also investigated the importance of the training data on the model ability to fit brain data, comparing models trained on Wikipedia and on the Full dataset (Wikipedia + Gutenberg xlarge). These analyses were performed on LSTM and GloVe.

\section{Results}
\label{results}

Figure~\ref{fig:rvalue-distribs} shows the distributions of R$_{\mbox{test}}$-values across voxels for trained and untrained models of various architectures and number of layers. 

\subsection{Performance of untrained models}

Remarkably, all untrained models, regardless of their architecture, explain signal better than chance. 
Untrained LSTM and untrained GloVe (that is, Random Embeddings) perform equally well with an average score around 6.3\% (SE=0.02\%), and significantly better than Transformers as attested by direct comparisons between 4 layers models: LSTM.4$-$GPT-2.4 (1.6\% SE=0.02\%); LSTM.4$-$BERT.4 (0.7\%  SE=0.004\%). Overall, untrained \GPT{}.4 had the worst performance (BERT.4$-$\GPT{}.4 (0.9\% SE=0.01\%)). 

\begin{figure}
    \centering
    \includegraphics[width=\columnwidth]{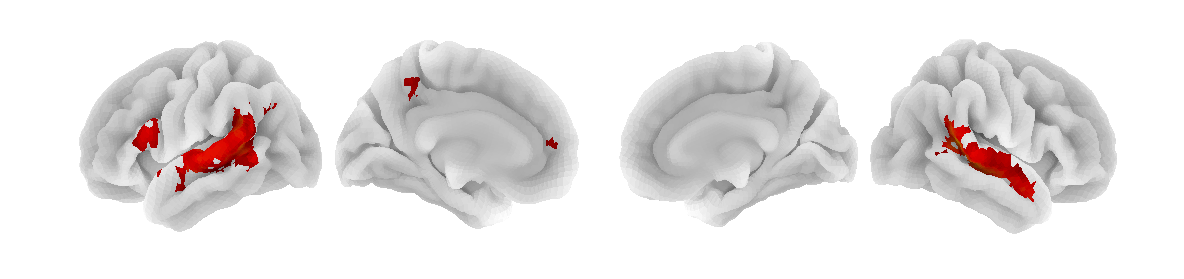}
    \includegraphics[width=.8\columnwidth]{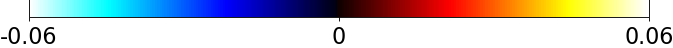}
    \caption{\textbf{LSTM vs. GPT-2 architecture:} Brain regions in which an untrained LSTM outperforms an untrained GPT-2 model. The comparison is for 2-layer models.}
    \label{fig:lstm_vs_gpt}
    
\vskip 0.2in
\begin{center}
\textsl{\small A. Increases in \rtest{} values with training}\\
\includegraphics[width=.8\columnwidth]{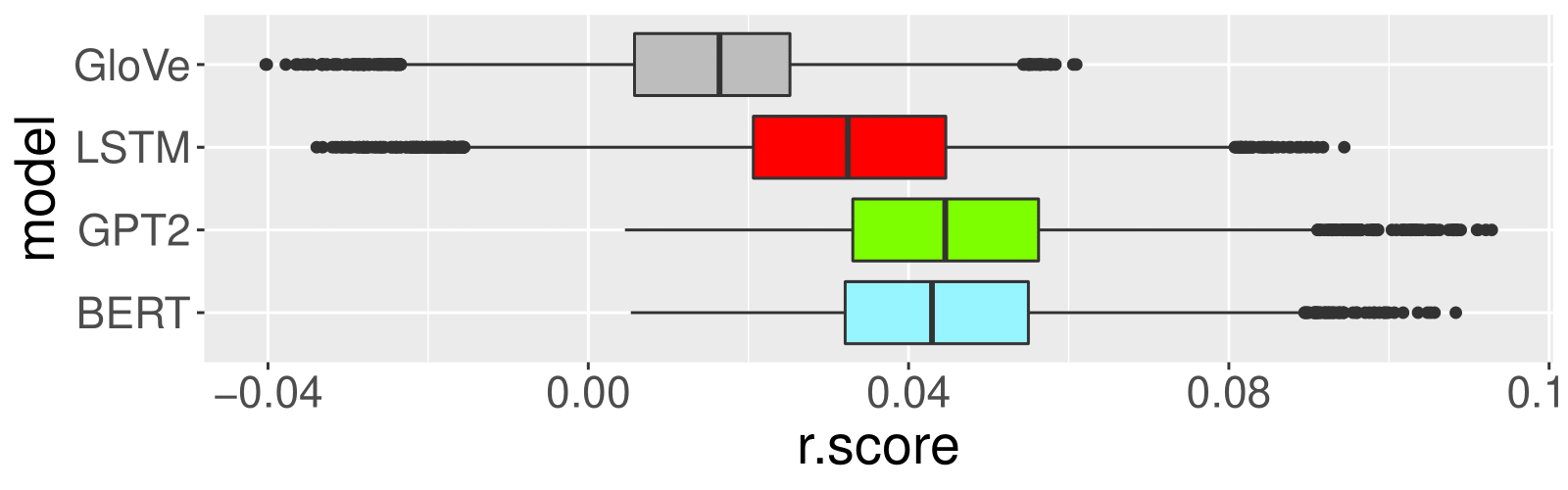}
\vskip 0.05in
\textsl{\small B1. LSTM$_{\mbox{trained}}$ - LSTM$_{\mbox{untrained}}$}\\
\centerline{\includegraphics[width=\columnwidth]{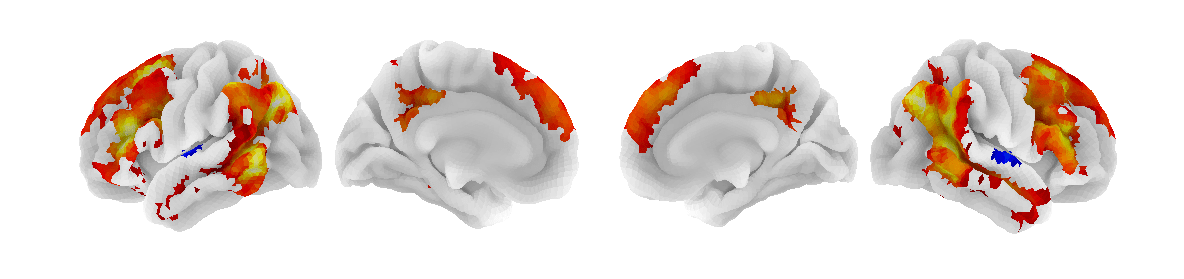}}
\vskip -0.15in
\textsl{\small B2. GPT-2$_{\mbox{trained}}$ - GPT-2$_{\mbox{untrained}}$}\\
\centerline{\includegraphics[width=\columnwidth]{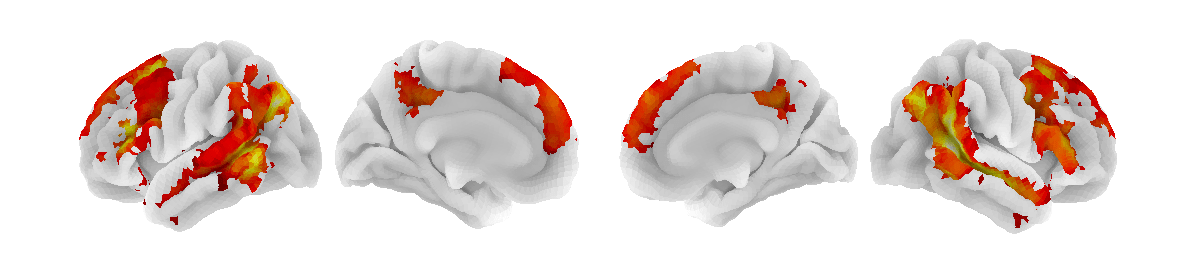}}
\vskip -0.15in
\textsl{\small B3. BERT$_{\mbox{trained}}$ - BERT$_{\mbox{untrained}}$}\\
\centerline{\includegraphics[width=\columnwidth]{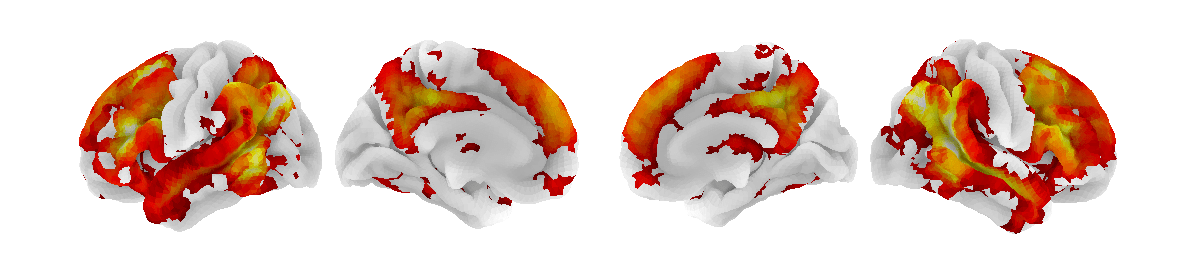}}
\centerline{\includegraphics[width=.8\columnwidth]{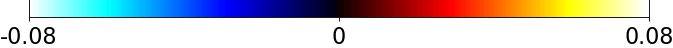}}
\vskip 0.01in
\textsl{\small C. Intersection of untrained models overlap and training gain overlap}\\
\centerline{\includegraphics[width=\columnwidth]{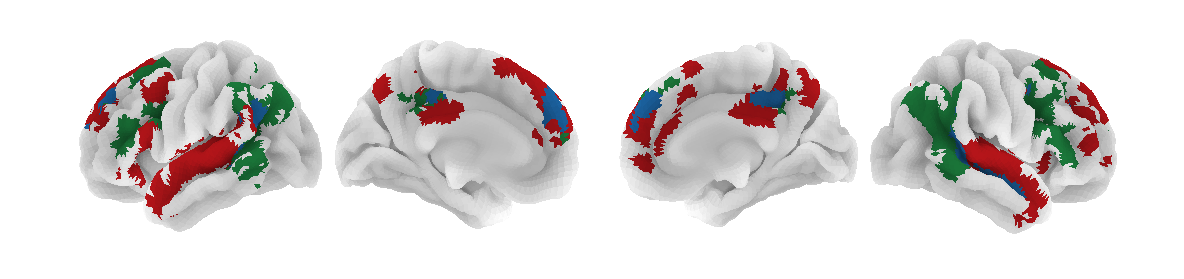}}
\centerline{\includegraphics[width=\columnwidth]{Figures/Fitting_brain_data/legend_overlap.pdf}}

\caption{\textbf{Effect of model training.} A) Distributions of  R$_{\mbox{test}}$ increase for the 2-layer versions of the three types of models. B) Brain areas showing significant increases: LSTM (top), \GPT{} (middle) and BERT (bottom). 
C) Regions showing the strongest gains in R scores with training across the three models (intersection of the three previous maps thresholded at the 10\% upper percentile), in green. Regions showing the strongest R scores across the three 2-layer untrained models: LSTM, \GPT{}, BERT (intersection of the three maps thresholded at the 10\% upper percentile), in red. There is a 18\% overlap between these two highlighted networks (in blue).}

\label{Fig:training_effect}
\end{center}
\end{figure}

The brain regions where LSTM$_{\mbox{untrained}}$ performs significantly better than \GPT{} are displayed on Fig.\ref{fig:lstm_vs_gpt}. They are located within the left hemispheric language network and its right counterpart (Superior Temporal Gyrus/Superior Temporal Sulcus and Inferior Frontal Gyrus (pars opercularis)). 
    
\subsection{Effect of number of layers}

Next, we looked at the effect of number of layers for LSTM, \GPT{} and BERT models.

As can be seen on Fig.~\ref{fig:rvalue-distribs} the change in performance for untrained models is either flat (for LSTM and BERT) or negative (for \GPT{}). Comparing 4-layer models to 1-layer models yields the following: LSTM (-0.02\%  SE=0.002\%); \GPT{} (-0.6\% SE=0.004\%), BERT (-0.02\% SE=0.003\%). 

For trained models, performance improves with the number of layers. The increase in performance (4-layer model's performance - 1-layer model's performance) is more marked for Transformers --- \GPT{} (2\% SE=0.006\%) and BERT (1\% SE=0.006\%) --- than for LSTM (0.4\% SE=0.005\%).

\subsection{Effect of Training}

Visual inspection of Fig.\ref{fig:rvalue-distribs} shows, unsurprisingly, that training helps: trained model fit brain data better than models initialized with random weights. To quantify this improvement for each model type, we computed, in each voxel, the difference in \rtest{} between the trained model and the untrained model. All differences were statistically significant: GloVe (1.5\% SE=0.02\%); LSTM (3.1\% SE=0.02\%) ; \GPT{} (4.5\% SE=0.02\%); BERT (4.4\% SE=0.02\%); in Student T-tests, all $p$s $< 10^{-16}$). Fig.\ref{Fig:training_effect}A shows the distributions of these training effects.

The maps on Fig.\ref{Fig:training_effect}B show the locations of voxels where the \rtest{} increases are significant.  The effect of training is spatially consistent across models, that is, displays similar topographies across models; and the R-score improvements are comparable in high-order language networks across models.

To assess the similarity between the hotspots on these maps, we thresholded them, keeping the 10\% of voxels (2617 voxels) showing the highest gains with training. We then computed the percentage of overlap across the resulting binarized maps. Results are presented in Table~\ref{tab:Overlap}. The overlap between the three maps is 75\% across all 3 models. 
We then thresholded the untrained models brain maps, keeping again the 10\% (2617 voxels) of voxels showing the highest r score, and computed the percentage of overlap across the resulting binarized maps. Results are presented in Appendix Table~\ref{tab:Overlap}. The overlap between the three maps is 79\% across all 3 models. 
The differences and similarities of these two overlaps are displayed in Fig.\ref{Fig:training_effect}C.

\begin{table}[b]
    \centering
    \begin{tabular}{ l c  r  }
     \hline
     \textbf{Model}  & GPT-2     & BERT  \\ 
     \hline
     LSTM            &  79\%    & 86\% \\ 
     GPT-2           &    .     & 85\% \\ 
     \hline
    \end{tabular}
    \caption{\textbf{Overlap between training effect brain maps} The percentage of common voxels when the maps were thresholded at their 10\% upper percentile.}
    \label{tab:Overlap}
\end{table}

\subsection{Comparisons between models}
\label{results:comparing_models}

\begin{figure}
\centering
\textsl{A. LSTM $-$ GloVe}
\includegraphics[width=\columnwidth]{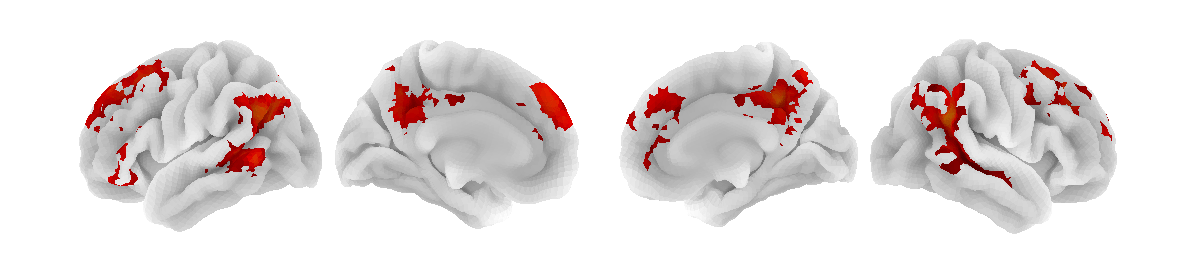}

\textsl{B1. BERT $-$ LSTM}
\includegraphics[width=\columnwidth]{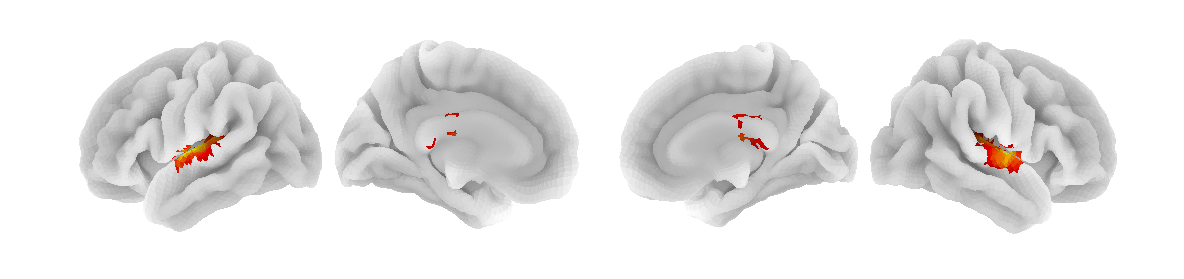}

\textsl{B2. BERT $-$ GPT-2}
\includegraphics[width=\columnwidth]{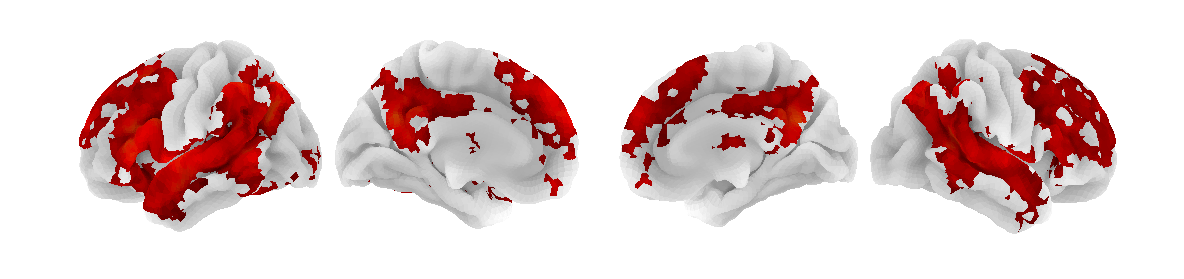}

\textsl{C1. (BERT.1 $-$ LSTM.1) $\times$ (Trained $-$ Untrained)}
\includegraphics[width=\columnwidth]{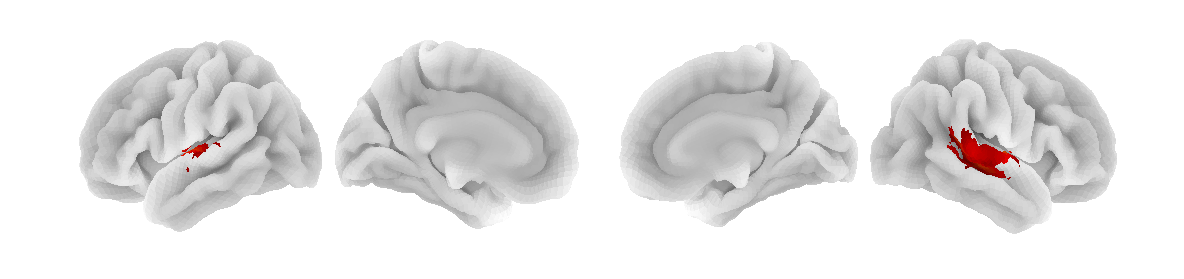}
\textsl{C2. (BERT.2 $-$ LSTM.2) $\times$ (Trained $-$ Untrained)}
\includegraphics[width=\columnwidth]{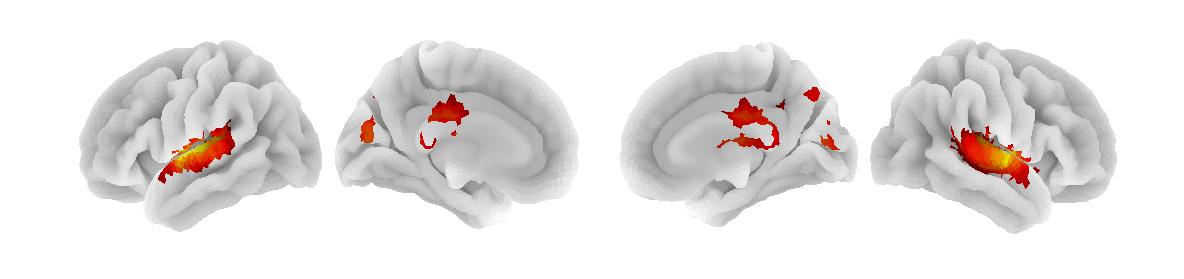}
\textsl{C3. (BERT.4 $-$ LSTM.4) $\times$ (Trained $-$ Untrained)}
\includegraphics[width=\columnwidth]{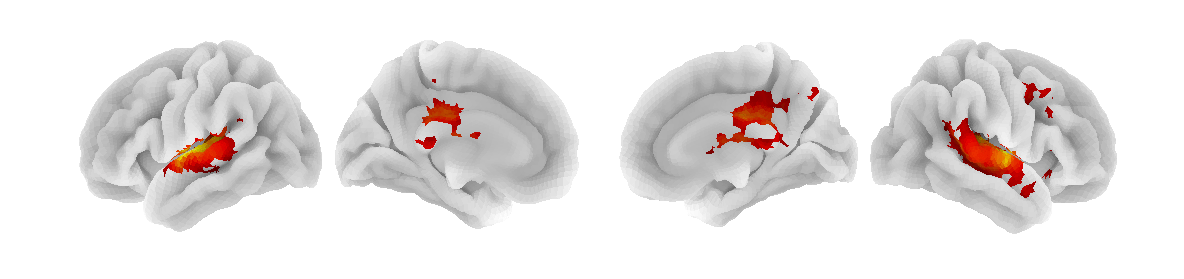}

\includegraphics[width=.8\columnwidth]{Figures/colorbars/colorbar_006.png}

\caption{\textbf{Localisation of differences between models} A) Comparison between a contextual model (LSTM) and a non-contextual model (GloVe). LSTM outperforms the non-contextual model GloVe in core regions of the language network, with significant effects that are bilateral and included in the core of the previously highlighted network of regions that are better fitted with training \ref{Fig:training_effect}C.
B1) Comparison of a transformer-based model (BERT) with a recurrent neural network (LSTM). Trained BERT better fits fMRI brain data than LSTM model around Heschl's Gyri bilaterally.
B2) Comparison of a bidirectional transformer-based model (BERT) with an incremental transformer-based model (\GPT{}). Trained BERT better fits fMRI brain data than \GPT{} in the entire language network.
C) Interaction between model architecture (BERT vs. LSTM) and Training (trained vs. untrained) for 1-layer (top), 2-layer (middle) and 4-layer (bottom) models. BERT benefits more from training than LSTM. The more layers, the more it learns. }
\label{Figure:comparing_models}
\end{figure}

Then, we ran a comprehensive comparison of the models described in Section~\ref{preliminaries}.
Firstly, we contrasted a model that takes context into account (LSTM) to a model that does not (Glove). Importantly, to conduct a fair comparison, both models were trained on exactly the same corpus, and had vocabulary of same size (see Section~\ref{preliminaries}). 
The contrast map displayed in Fig.~\ref{Figure:comparing_models}A highlights a set of regions located in the temporo-parietal junction, similar to the trained models overlap of Fig.\ref{Fig:training_effect} (in green), with bilateral effects in the medial and lateral Superior Frontal Gyri and Posterior Cyngulate gyri, as well as left-lateralized effects in the Temporal Pole, the Inferior Frontal and Middle Temporal gyri.

Note that this map is quite similar to the green network showing an effect of training across models (Fig.\ref{Fig:training_effect}C). This suggests the role of these regions in context processing, given that in both cases, the model benefits from using context for predicting activity in these regions.

Next, we compared  a model using attention (BERT) to a recurrence-based model (LSTM). Fig.~\ref{Figure:comparing_models}B1) shows that BERT outperforms LSTM mostly in the Superior Temporal Gyri, and in the auditory cortex. 

Comparing the bidirectional BERT with the incremental \GPT{} in Fig.\ref{Figure:comparing_models} B2) shows that BERT outperforms \GPT{} in the entire language network.
Finally, to test which model architecture best benefit from training, we investigated the interaction between model architecture (BERT vs. LSTM) and training (trained vs. untrained). The interaction map is shown on Fig.~\ref{Figure:comparing_models}C. The effects are more spread than the direct difference between the trained BERT and LSTM, as expected from the fact that LSTM untrained has higher performance than BERT untrained. 
Transformer-based models gain more with training relative to LSTM (this is also the case for GPT2, see Appendix Fig.S6C), and explorations show that this relative gain increases with the number of layers. 

In summary, \textit{i)} LSTM model outperforms the non-contextual model GloVe in core regions of the language network; \textit{ii)} Transformer-based models benefit more from training than the LSTM model, and achieve higher brain scores (compared to LSTM) mostly around the auditory cortex. 

\subsection{Relationship between Perplexity and Brain score}
\label{results:perplexity_and_brain_score}

\begin{figure}[ht]
    \centering
    \includegraphics[width=\columnwidth]{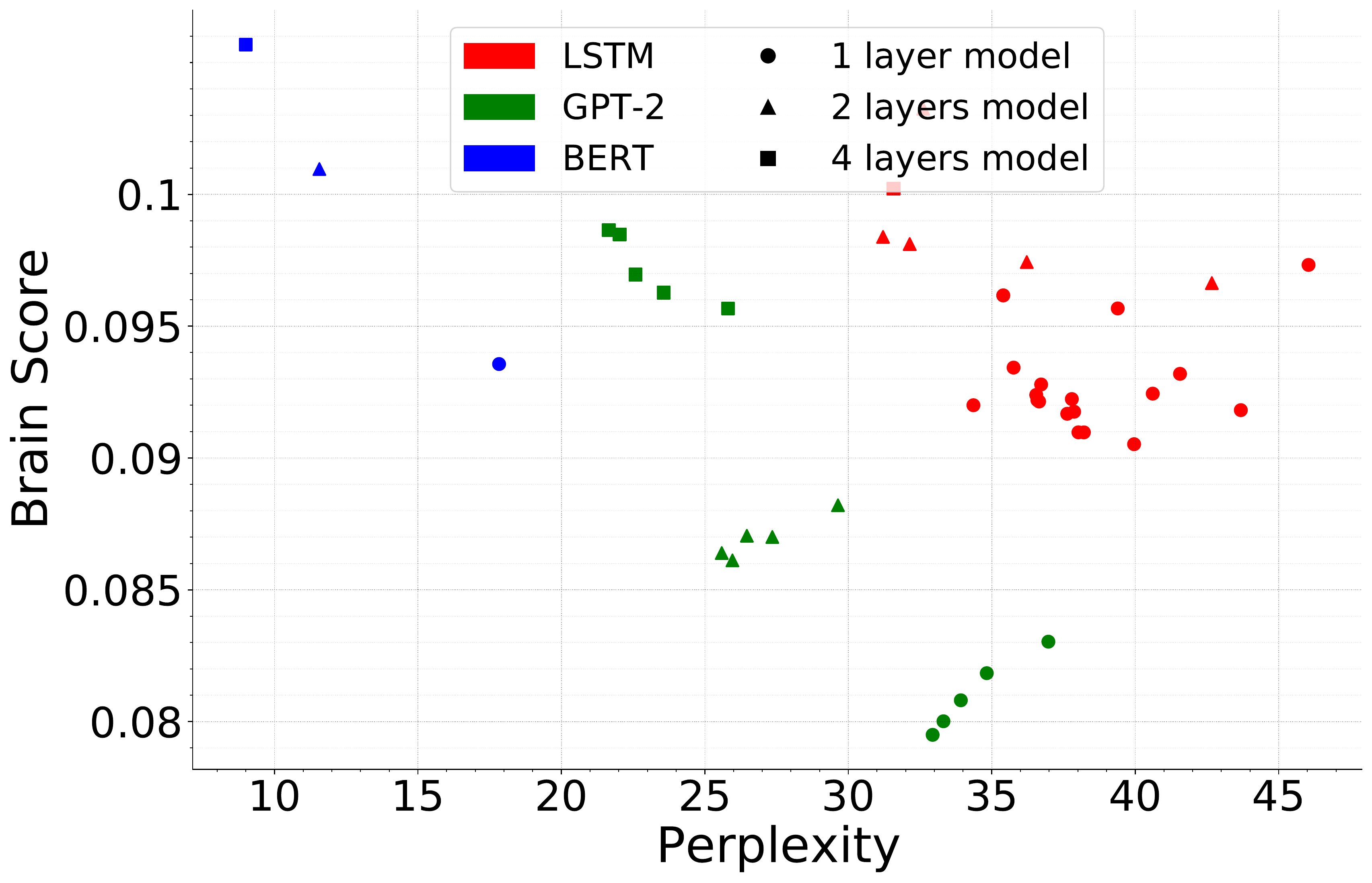}
     \caption{\textbf{Perplexity does not predict Brain score.} Among all instances of LSTM, GPT-2 and BERT that are plotted, we found no monotonic relationships, showing that Perplexity cannot serve as a simple proxy to determine Brain Score, as the relation between the two is impacted by the model class, its architecture and training. We represented the brain scores and perplexities of BERT (blue), of several instances of GTP-2 at different training stages (green) and of various instances of LSTM trained on different datasets (red).
     }
    \label{Figure:perplexity_vs_brain_score}
\end{figure}

Fig. \ref{Figure:perplexity_vs_brain_score}A shows the relationship between perplexities (model loss) and brain scores derived from various models, architectures, training sets and training stages. Unlike previous reports \citep{schrimpf_artificial_2020}, we did not observe a clear monotonic relationship between the two variables. For example, the average LSTMs perplexity is worse than that of GPT-2, but the average brain score is higher.

We investigated in more details the effects of model class, number of layers, training epochs and training dataset size on the relationship between brain score and perplexity. The results are presented in Appendix Fig.\ref{appendix:Figure:perplexity_vs_brain_score}. 
In Fig.S\ref{appendix:Figure:perplexity_vs_brain_score} panel A, we observed that within each model class, increasing the number of layers improves perplexity and brain score.
However, within a given model class, there is not always a monotonic relationship between brain score and perplexity as shown by the effect of training epochs in panels B and C for GPT-2 and panel D for LSTM.
Finally, manipulating training dataset size with LSTM shows no simple relation between brain score and perplexity.

\subsection{Effect of Training set}

\begin{figure}[ht]
    \centering
     \textsl{A. LSTM(full) $-$ LSTM(Wikipedia)}
     \includegraphics[width=\columnwidth]{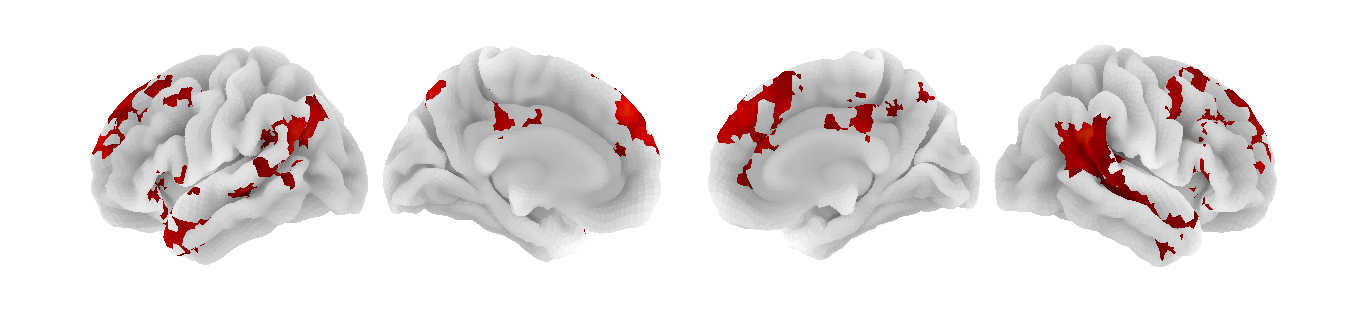}
     \textsl{B. GloVe(full) $-$ GloVe(Wikipedia)}
     \includegraphics[width=\columnwidth]{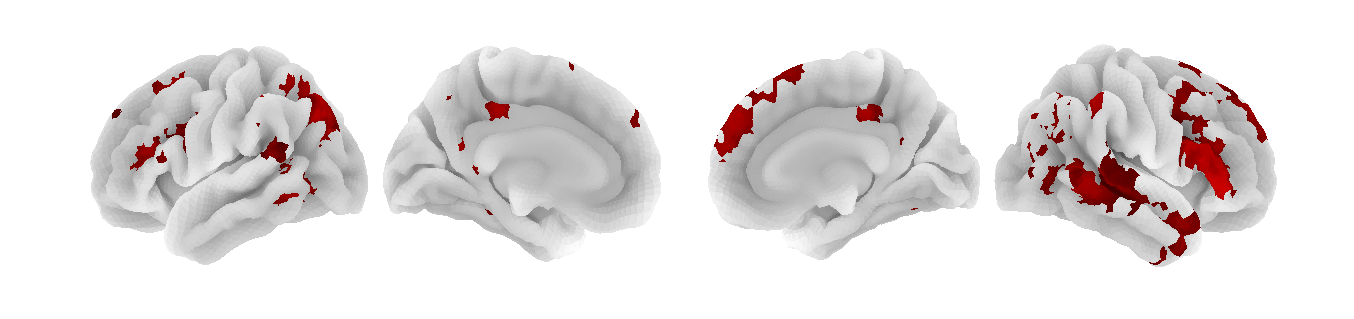}
     \centerline{\includegraphics[width=.8\columnwidth]{Figures/colorbars/colorbar_006.png}}
    \caption{\textbf{Influence of Training dataset on  $R_{test}$.} LSTM and GloVe better fit brain data when learning on more training data. This shows the dependence of models contrast on training data. Our full dataset comprised Gutenberg + Wikipedia (4.8 GB) while Wikipedia represented 425 MB.}
    \label{fig:training}
\end{figure}

Data used for training have a strong influence on the outcome. 
Fig.\ref{fig:training} presents contrasts maps obtained when training LSTM or GloVe with our custom \textit{Full} dataset versus Wikipedia. This shows that off-the-shelf models trained on Wikipedia likely lack statistical power to detect brain activations. 

\clearpage

\section{Discussion}
\label{discussion}

Previous work has shown that brain activity during visual or language processing can be significantly predicted from artificial neural-network activations. From a neuro-scientific point of view, the interest lies in the possibility to study language processing in the brain by manipulating the information provided to an artificial model and then to assess the impact on the model's ability to fit brain data. 
In the present research, we examined the impact of several factors on the fitting performance of artificial models. We studied the impact of model architecture (assessing GloVe and LSTM, \GPT{} and BERT models with varying number of layers), models' perplexity and training corpus, on their capacity to predict functional Magnetic Resonance Imaging timecourses of participants listening to an audiobook. We made several observations: (1) untrained versions of each model already explain significant amount of signal in the brain, with the untrained LSTM and GloVe outperforming the others; (2) training NLP models improves brain scores in the same set of brain regions irrespective of model’s architecture; (3) Perplexity is not a good predictor of brain score; (4) training data have a strong influence on the capacity to fit brain data. 

One discovery is that all architectures are not equal, but training them consistently increases brain scores in the same set of brain areas. Moreover, these very brain regions are also better predicted by models that take the context into account, such as LSTMs, compared to static models, represented here by GloVe. This provides converging evidence that these regions perform context-dependent computations. From a neuratomical point of view, they are located on the border of the core language regions (IFG and STS) and partly overlap with regions assigned to the Default mode network (Angular Gyri, Dorso mesial prefrontal cortex). These results are coherent with previous work investigating the processing of contextual information by either using LSTM models \citep{jain_incorporating_2018} or by scrambling the stimuli at different levels \citep{lerner_topographic_2011}, confirming that this network (in green/blue in Fig.\ref{Fig:training_effect}) is at the center of combinatorial language processing in the human brain. 

We observed that even if transformers start with a disadvantage regarding the ability to fit fMRI brain data, they benefit more from training than LSTMs, and they are able to take advantage of stacks of layers to improve their fitting performance. The comparison of untrained LSTM and untrained GloVe (i.e., random embeddings) showed no significant differences, whereas the comparison of untrained GloVe and untrained transformers showed significant R-score differences in some regions. The difference between untrained LSTMs and untrained Transformers might be due to their different architectures. However, there might be an alternative explanation. Note that for untrained GloVe (random embeddings), each word in the corpus is assigned a \textit{fixed} vector, whereas for untrained Transformers, each word is mapped to a \textit{variable} vector, depending on the context that surrounds the word. Therefore, untrained Glove might better predict brain responses to words that occur frequently both in the training and in the test data (e.g., function words). In contrast, untrained transformers might generate very different embeddings to the same word (e.g., 'the' in the train and test sets), due to their context sensitivity. This variability could therefore reduce the brain score of untrained Transformers compared to that of untrained GloVe. Finally, our results suggest that untrained LSTM are more similar to untrained Glove, having less context sensitivity compared to Transformers. Taken together, this suggests that most of what the untrained baselines capture is similarity in brain responses to words that appear in both the train and test sets.

The discrepancy between brain score and perplexity indicates that training is not a guarantee of convergence towards brain-like representations (see also \citealp{hale2019text}). Relatedly, other research also seems to militate against perplexity as a royal road to cognitive models (see, e.g.,  \citealt{clark2000mindware} , chapter 11, 2$^{nd}$ edition).

A final methodological word of caution stem from our results: data used for training have a strong influence on the outcome, showing that off-the-shelf models trained on small datasets like Wikipedia lack statistical power to capture brain activations and should be avoided to probe brain representations.
%
%
%

\section*{Acknowledgements}
\label{acknowledgements}

This project/research has received funding from the American National Science Foundation under Grant Number 1607441 (USA), the French National Research Agency (ANR) under grant ANR-14-CERA-0001, the European Union’s Horizon 2020 Framework Programme for Research and Innovation under the Specific Grant Agreement No. 945539 (Human Brain Project SGA3), and the KARAIB AI chair (ANR-20-CHIA-0025-01). 


\bibliography{alex, yair}
\bibliographystyle{icml2022}

\newpage

\appendix

\onecolumn

\section{Convergence of the Language Models During Training}

\setcounter{figure}{0}
\begin{figure}[!ht]
\begin{center}
\centerline{
\includegraphics[width=0.33\columnwidth]{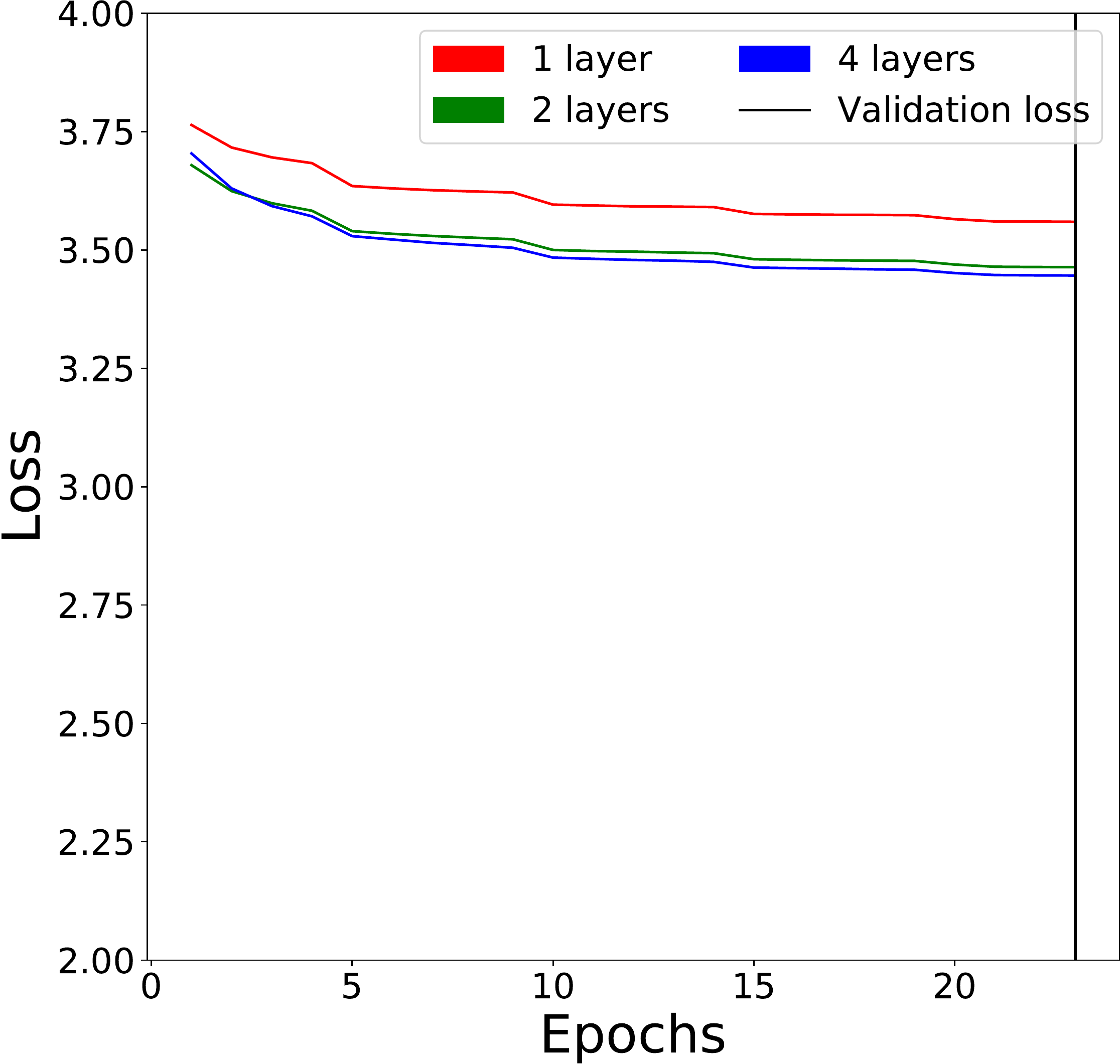}
\includegraphics[width=0.33\columnwidth]{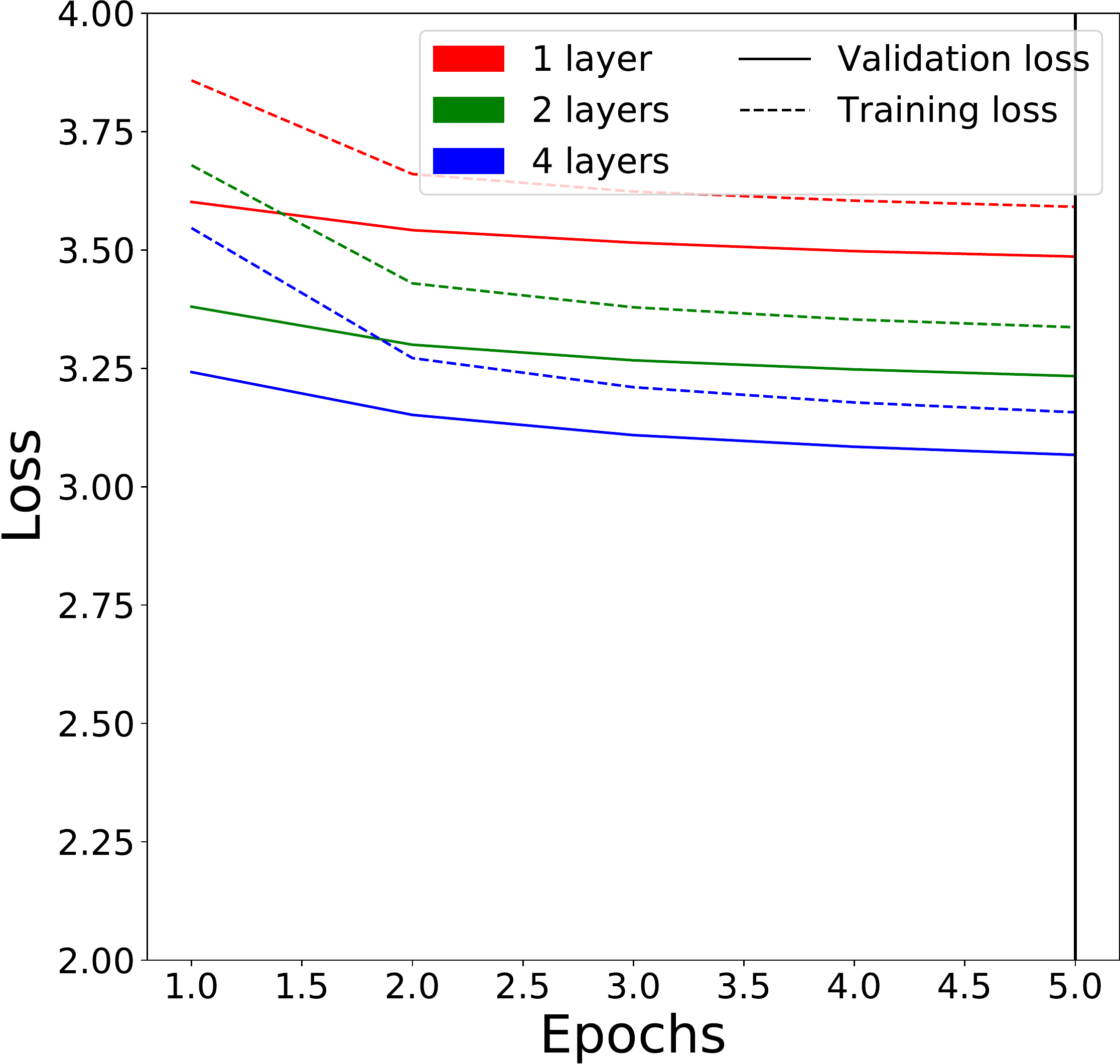}
\includegraphics[width=0.33\columnwidth]{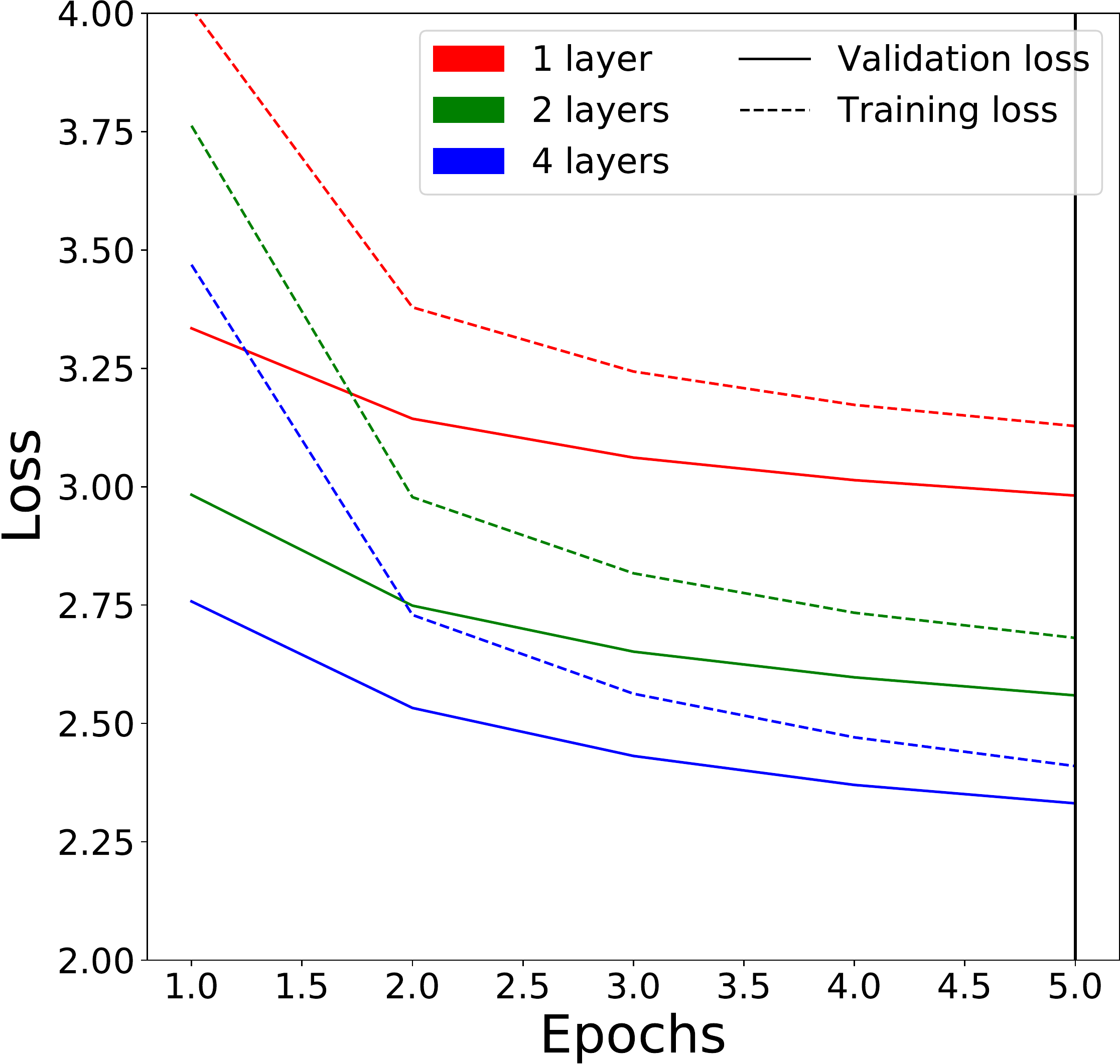}}

\caption{
\textbf{Model convergence during training on the Full dataset.} Validation losses for all trained models, as well as training losses for transformers, for LSTM (left), GPT-2 (middle) and BERT (right). LSTMs and Transformer-based models were trained for 23 and 5 epochs, respectively.}
\label{appendix:Figure:model_convergence}
\end{center}
\vskip -0.2in
\end{figure}

\begin{figure}[!ht]
\begin{center}

\textsf{A. BERT.2$_{\mbox{trained}}$ - BERT.2$_{\mbox{GOOGLE}}$}
\centerline{\includegraphics[width=.7\columnwidth]{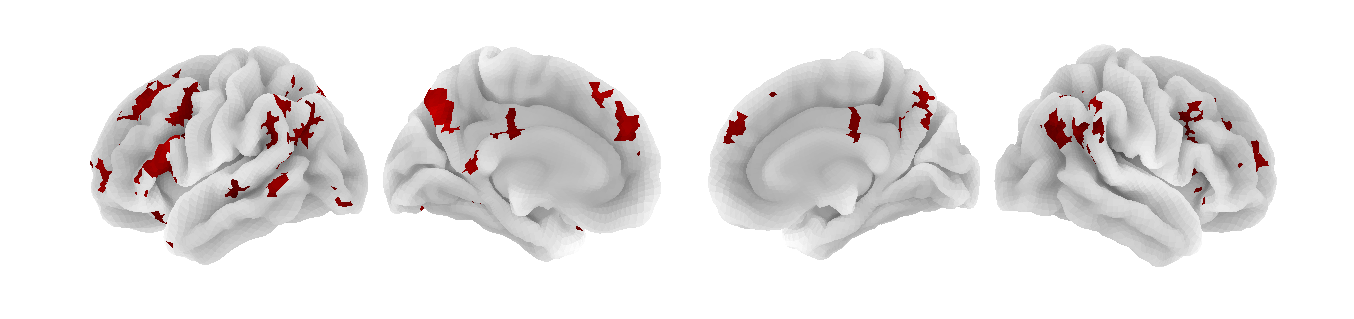}}
\textsf{B. BERT.4$_{\mbox{trained}}$ - BERT.4$_{\mbox{GOOGLE}}$}
\centerline{\includegraphics[width=.7\columnwidth]{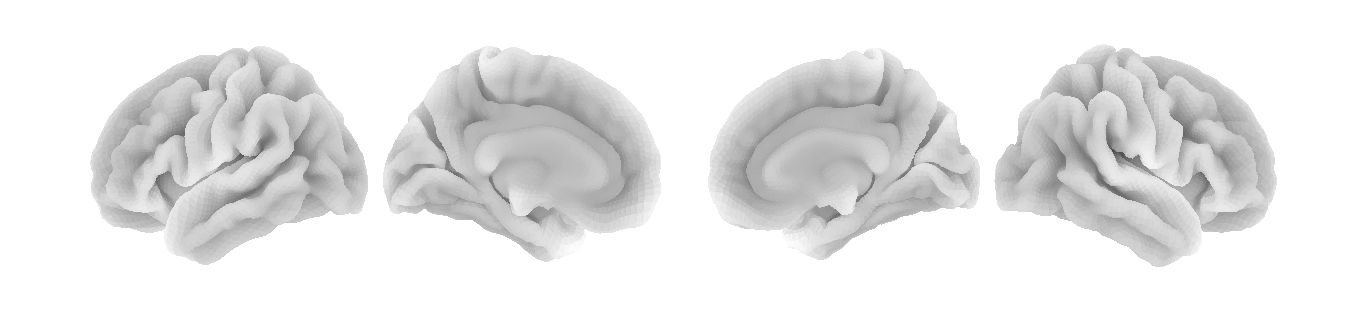}}
\textsf{C. BERT.4$_{\mbox{trained}}$ - BERT.12$_{\mbox{GOOGLE}}$}
\centerline{\includegraphics[width=.7\columnwidth]{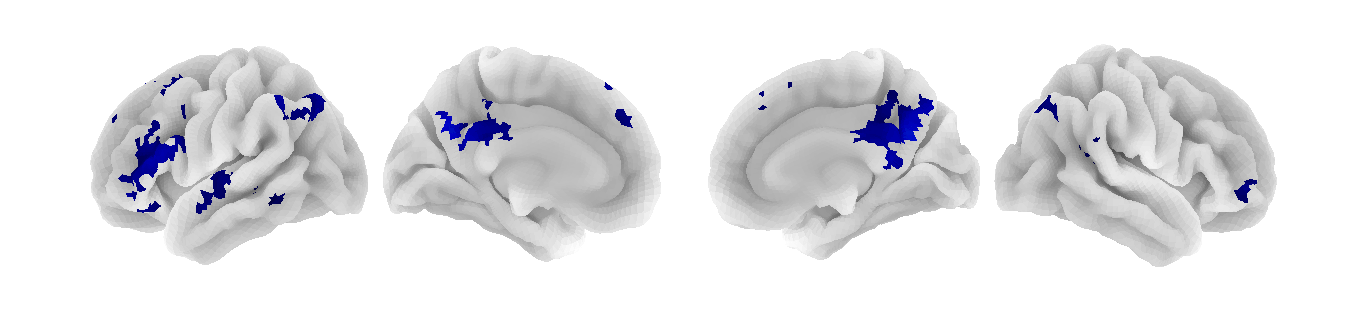}}
\centerline{\includegraphics[width=.5\columnwidth]{Figures/colorbars/colorbar_006.png}}

\caption{
\textbf{Comparison of the trained BERT models with off-the-shelf baselines.} To assess the performance of our trained models, we compared their ability to predict brain data with that of off-the-shelf models (
\url{https://github.com/google-research/bert}). The 2 and 4-layers BERT models either significantly outperform the baseline or are on a par. The 12-layers baseline, which is 3-times bigger than the 4-layers model, outperforms the latter in core regions of the language network, but only to a small extent.}
\label{appendix:Figure:comparison_with_baselines}
\end{center}
\vskip -0.2in
\end{figure}

\clearpage

\section{Evaluation of Brain-Fit Performance of the Untrained Models}

\begin{center}
\begin{tabular}{ l  c  c  r  }
 \hline
 \textbf{Model}    & LSTM  & GPT-2     & BERT  \\ \cline{2-4}
 LSTM               &   .  &  81\%    & 92\% \\ 
 GPT-2              &   .  &    .     & 86\% \\ 
 \hline
\end{tabular}    
\end{center}

\emph{Table 2. }\textbf{Overlap between untrained brain maps} The percentage of common voxels when the maps were thresholded at their 10\% upper percentile.
\label{SMtab:Overlap}

\begin{figure}[ht]
\vskip 0.2in
\centering
\textsf{A. Fitting brain data with LSTM.1$_{\mbox{untrained}}$}

\includegraphics[width=.7\columnwidth]{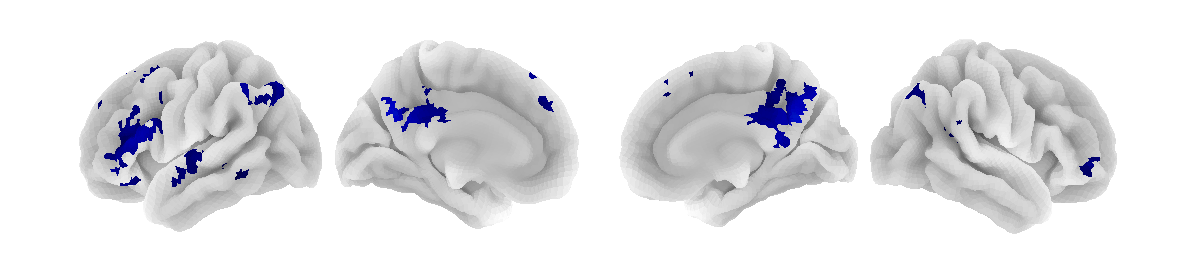}

\includegraphics[width=.5\columnwidth]{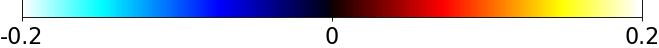}

\textsf{B. Untrained models overlap}

\includegraphics[width=.7\columnwidth]{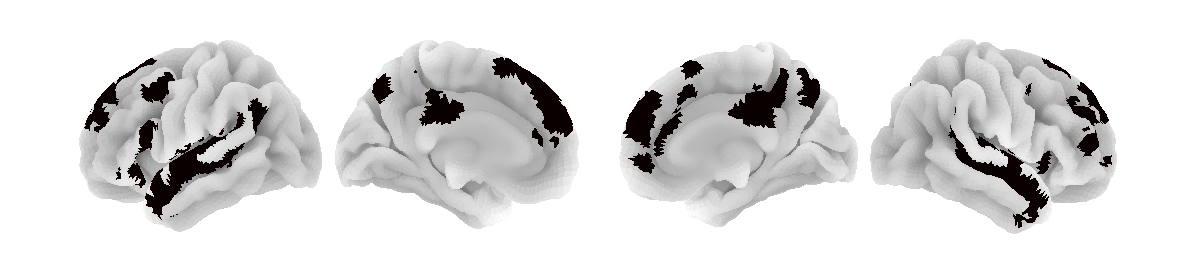}

\textsf{C. Training gain overlap}

\includegraphics[width=.7\columnwidth]{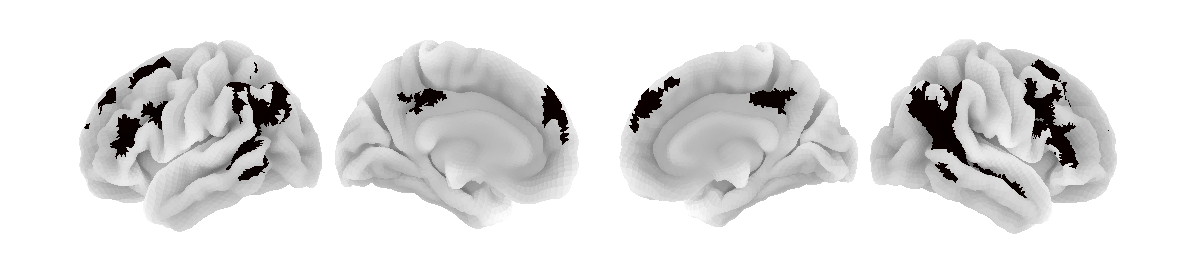}

\textsf{D. Intersection of untrained models overlap and training gain overlap}

\includegraphics[width=.7\columnwidth]{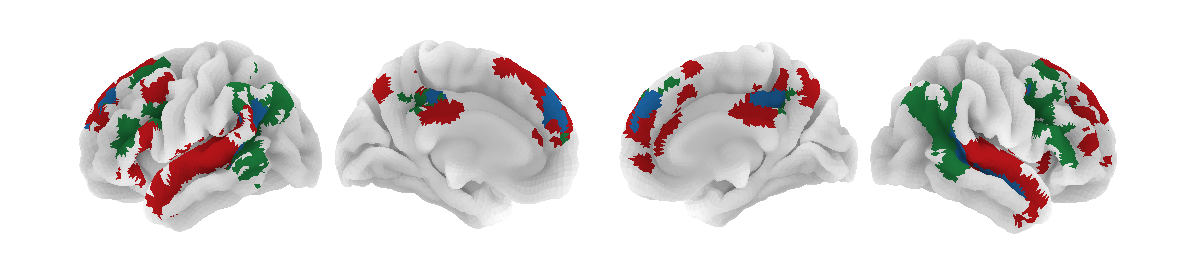} \\
\includegraphics[width=.7\columnwidth]{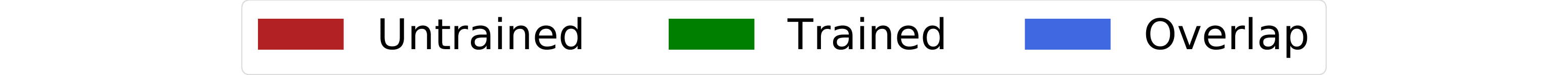}

\caption{
\textbf{A. Untrained LSTMs predict fMRI brain data better than chance across the entire brain.} Significant scores are observed in the language network, and non-significant scores in the motor cortex and the medial temporal regions.
\textbf{B. Regions showing the strongest R score across the three 2-layer untrained models: LSTM, GPT-2, BERT.} (intersection of the three maps thresholded at the 10\% upper percentile). There is a 79\% overlap across the three untrained models. \textbf{C. Regions showing the strongest gains after training: LSTM, GPT-2, BERT.} (intersection of the three maps thresholded at the 10\% upper percentile). There is a 75\% overlap across the three trained models. 
\textbf{D. Representing shared and specific brain regions of the two overlaps.} The regions showing the strongest R scores across the three untrained models only have a 18\% overlap with the regions showing the strongest gains across the three trained models.
}
\label{appendix:Figure:untrained}
\end{figure}

\clearpage

\section{Improvement in Brain Score after Training}

\begin{figure}[ht]

\centering
\textsl{A1. LSTM.1$_{\mbox{trained}}$ - LSTM.1$_{\mbox{untrained}}$}
\centerline{\includegraphics[width=.5\columnwidth]{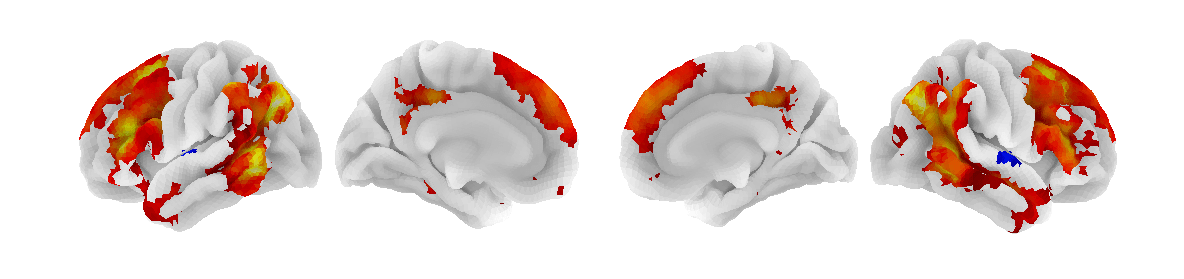}}
\vskip -0.1in
\textsl{A2. GPT-2.1$_{\mbox{trained}}$ - GPT-2.1$_{\mbox{untrained}}$}
\centerline{\includegraphics[width=.5\columnwidth]{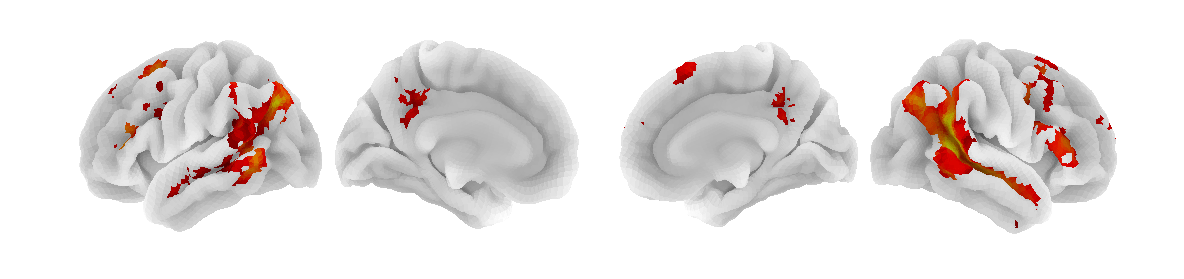}}
\vskip -0.1in
\textsl{A3. BERT.1$_{\mbox{trained}}$ - BERT.1$_{\mbox{untrained}}$}
\centerline{\includegraphics[width=.5\columnwidth]{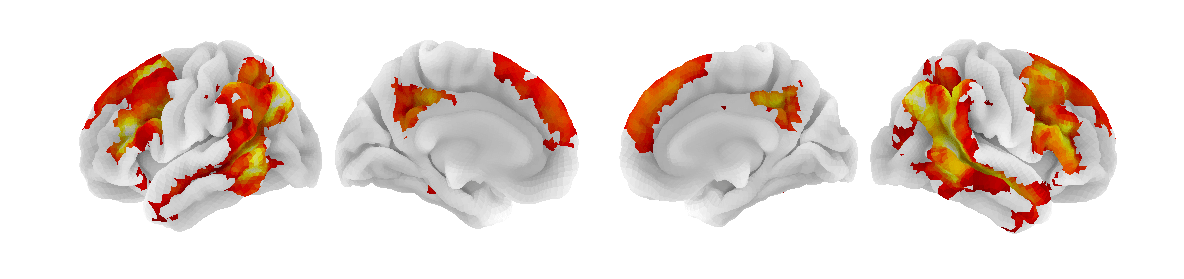}}
\vskip 0.1in
\textsl{B1. LSTM.4$_{\mbox{trained}}$ - LSTM.4$_{\mbox{untrained}}$}
\centerline{\includegraphics[width=.5\columnwidth]{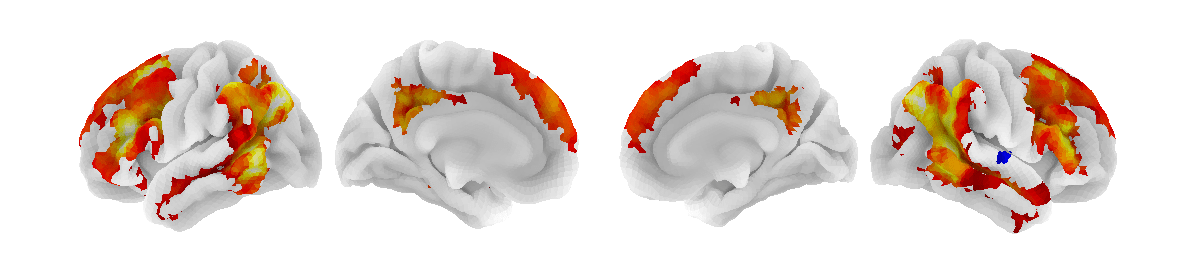}}
\vskip -0.1in
\textsl{B2. GPT-2.4$_{\mbox{trained}}$ - GPT-2.4$_{\mbox{untrained}}$}
\centerline{\includegraphics[width=.5\columnwidth]{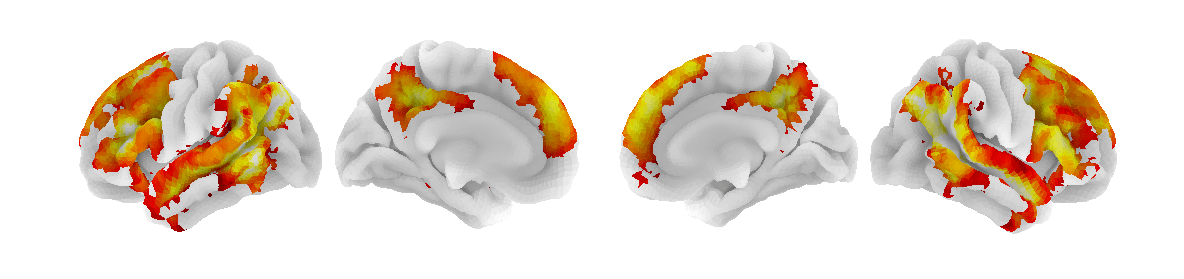}}
\vskip -0.1in
\textsl{B3. BERT.4$_{\mbox{trained}}$ - BERT.4$_{\mbox{untrained}}$}
\centerline{\includegraphics[width=.5\columnwidth]{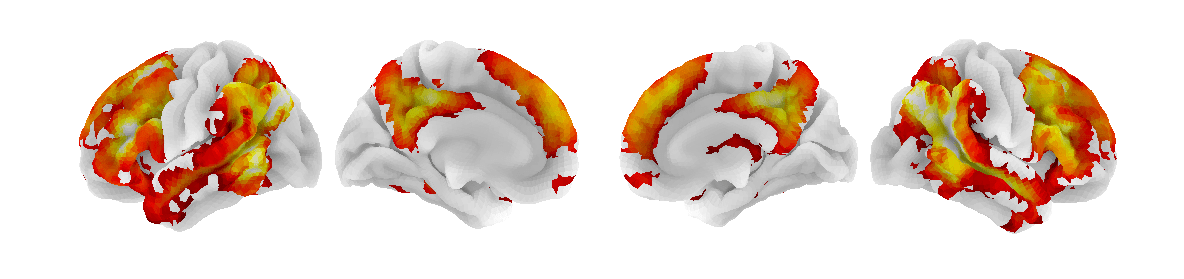}}
\centerline{\includegraphics[width=.4\columnwidth]{Figures/colorbars/colorbar_006.png}}

\caption{\textbf{A consistent increase in R scores due to training across various types of neural language models.} Contrasts between R scores of trained vs. untrained models. Contrast maps are shown for 1-layer models (panel A) and 4-layer models (panel B). In each panel, from top to bottom: LSTM, GPT-2, BERT.}
\label{appendix:Figure:trained}
\end{figure}


\clearpage
\section{Perplexity is not a Good Predictor of Brain Score}

\begin{figure}[ht]
\begin{center}

\begin{tabular}{p{0.45\textwidth}p{0.45\textwidth}}
\textsf{A. Effect of model class and number of layers} &  \\
\centerline{\includegraphics[width=7.5cm, height=5.5cm]{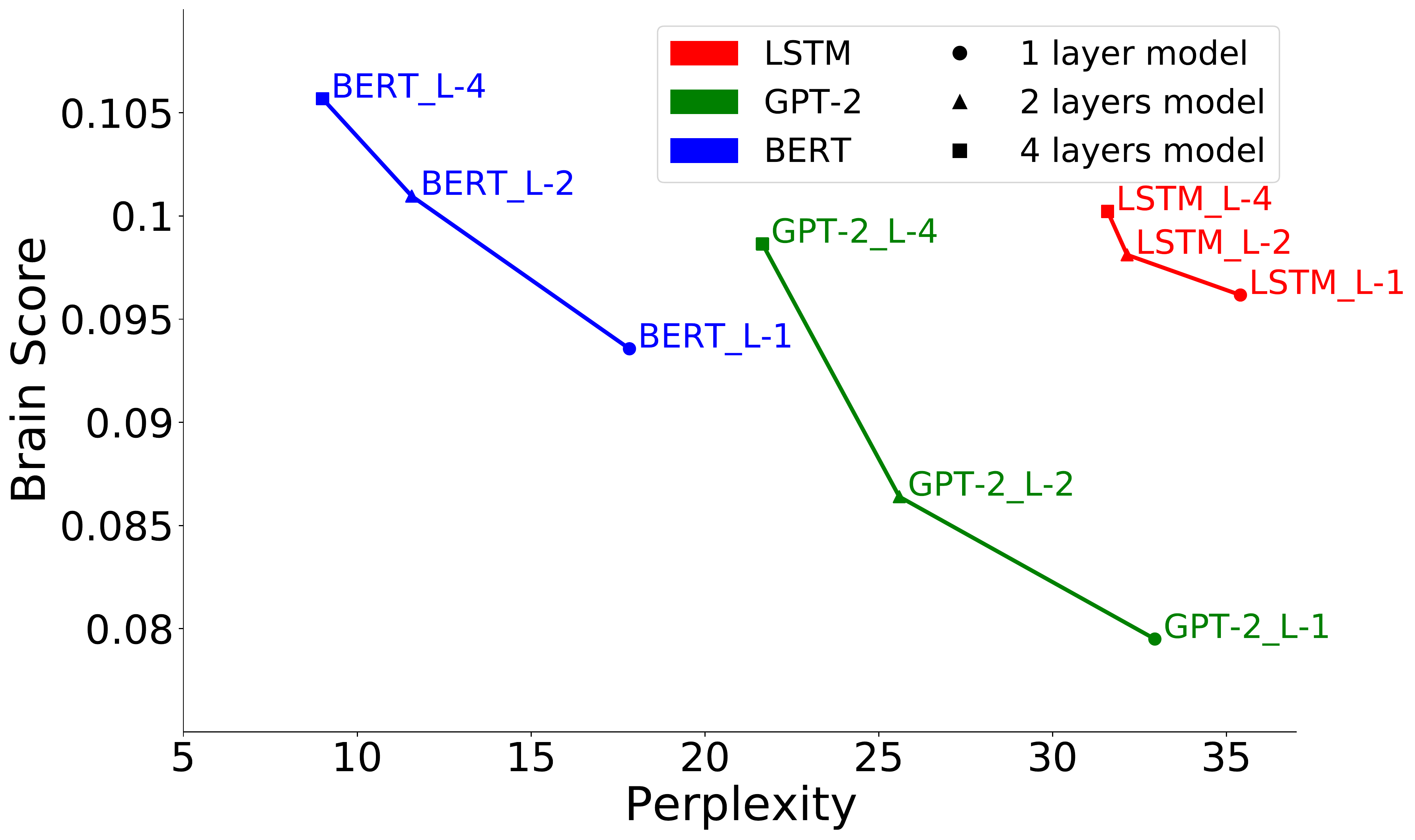}} &  
\\
\textsf{B. Effect of training epochs (GPT-2)} & \textsf{C. Effect of training epochs (GPT-2, Perplexity computed on TLP)} \\
\centerline{\includegraphics[width=7.5cm, height=5.5cm]{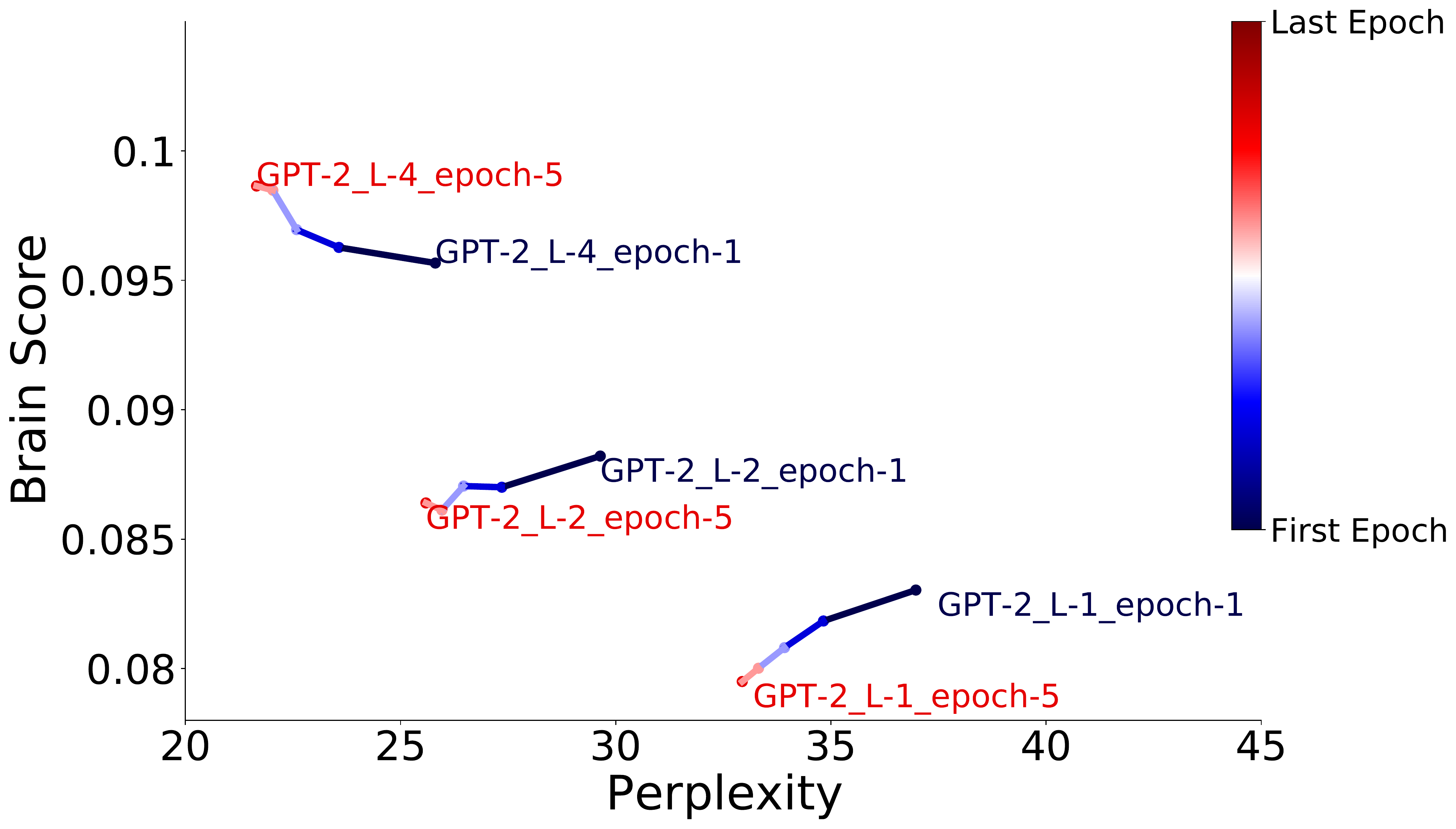}} & \centerline{\includegraphics[width=7.5cm, height=5.5cm]{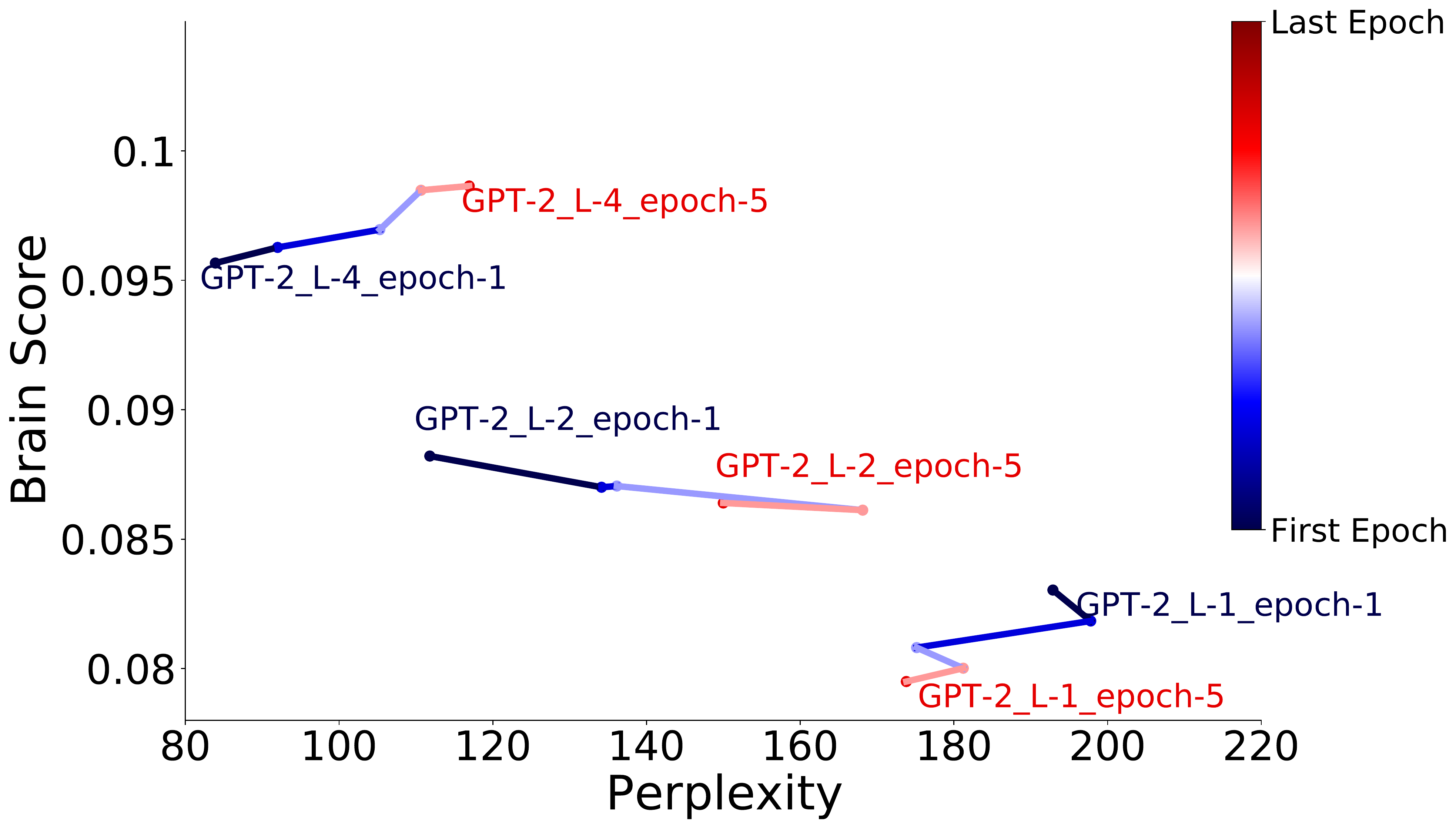}} \\

\textsf{D. Effect of training epochs (LSTM)}  & \textsf{E. Effect of training data size} \\
\centerline{\includegraphics[width=7.5cm, height=5.5cm]{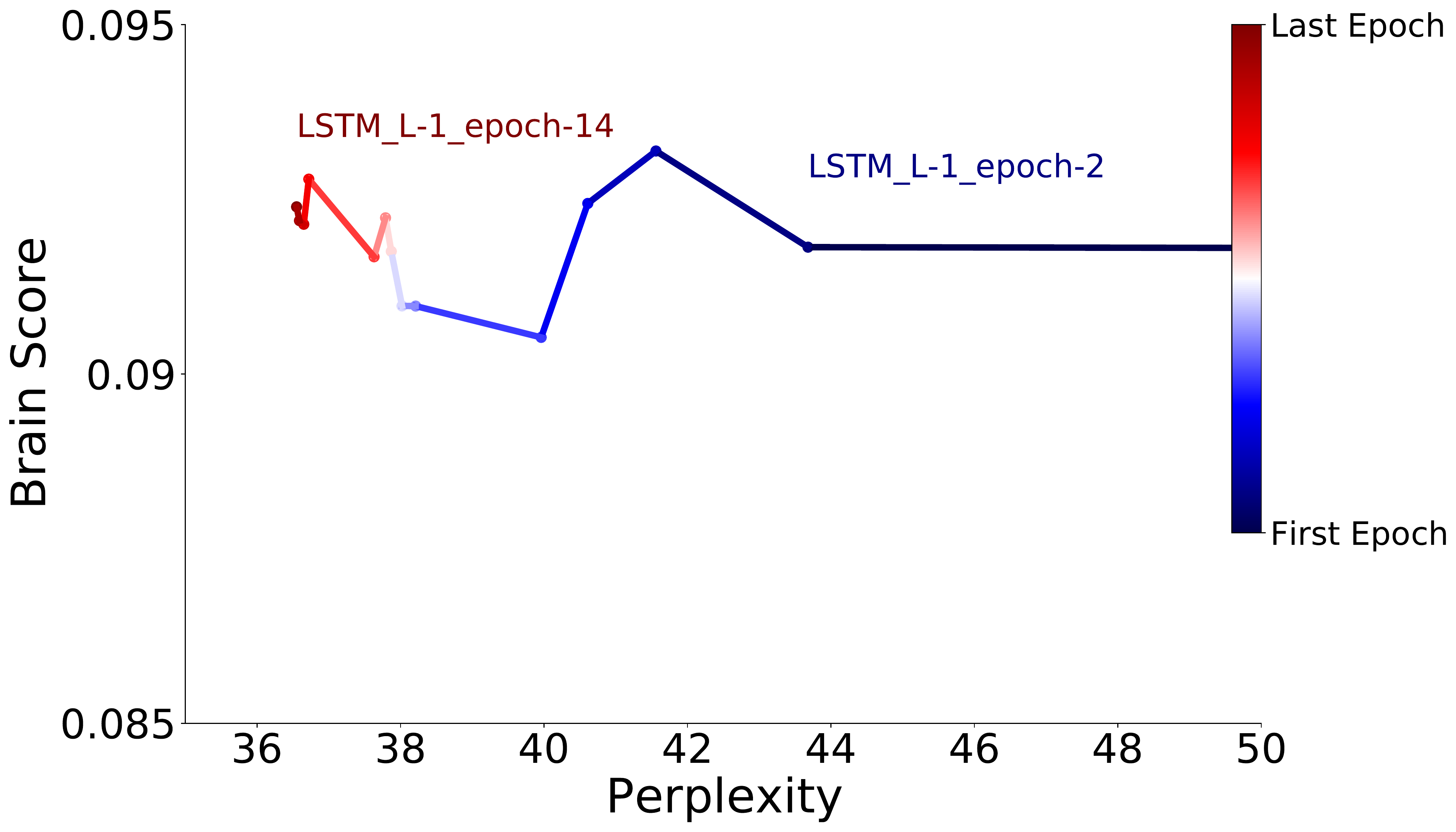}}& \centerline{\includegraphics[width=7.5cm, height=5.5cm]{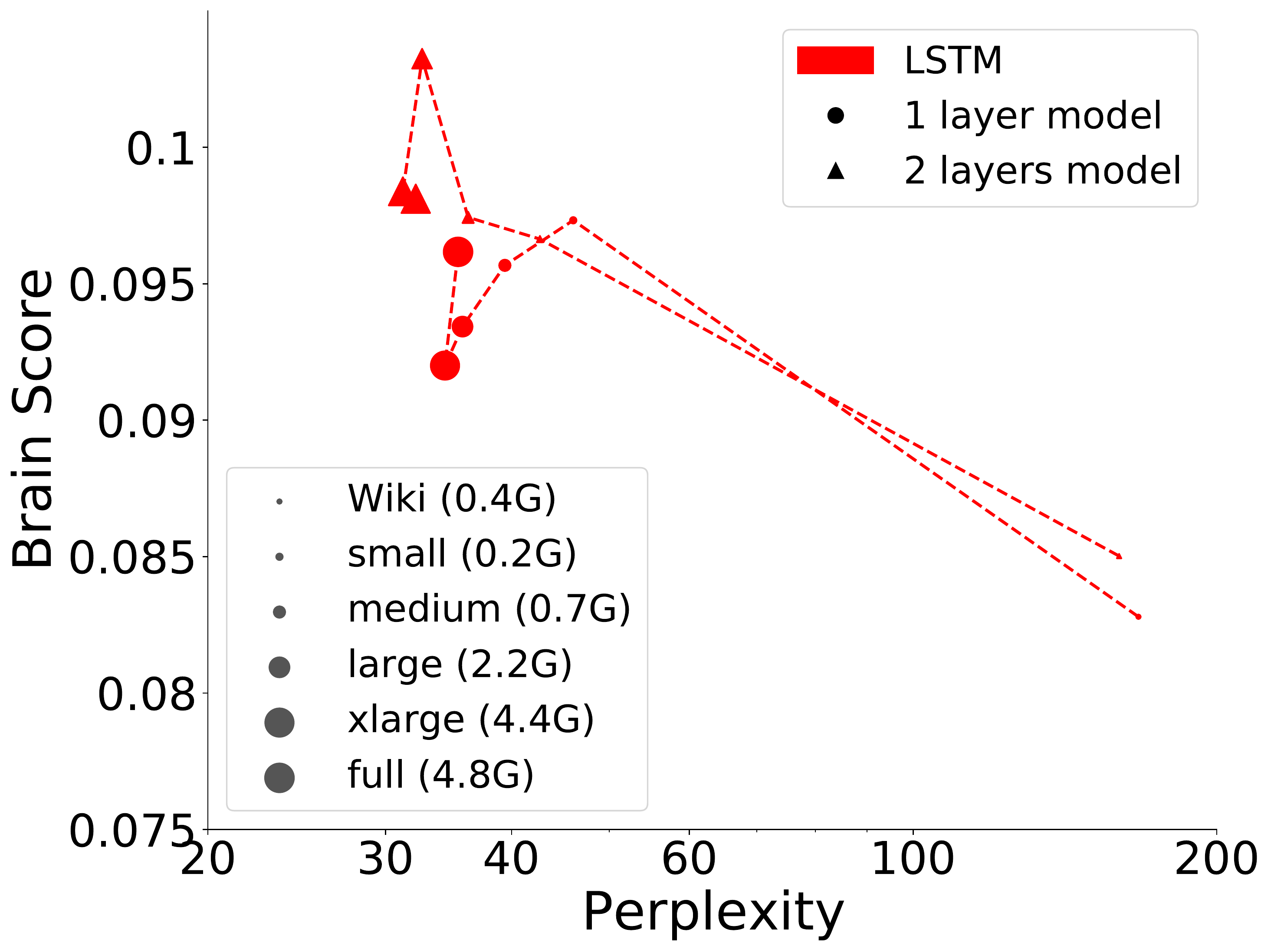}}

\end{tabular}

\caption{\textbf{Detailed analyses of the relation between brain score and perplexity as a function of model class (A), number of layers (A), training epochs (B-D) and training datasets (E).}
}
\label{appendix:Figure:perplexity_vs_brain_score}
\end{center}
\end{figure}

\begin{figure}[ht]

\begin{center}
\textsf{A. LSTM$_{\mbox{wikipedia}}$ - GloVe$_{\mbox{wikipedia}}$}
\centerline{\includegraphics[width=.5\columnwidth]{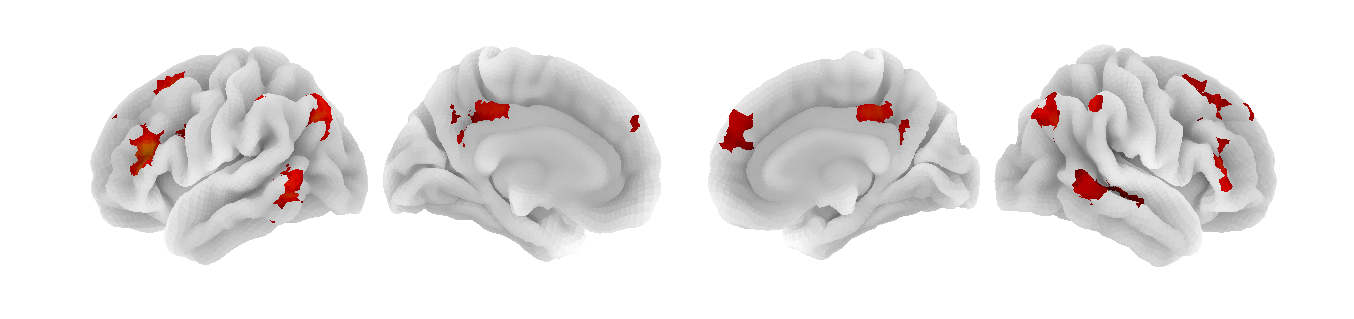}}
\vskip -0.05in
\textsf{B1. GPT-2.1 - LSTM.1}
\centerline{\includegraphics[width=.5\columnwidth]{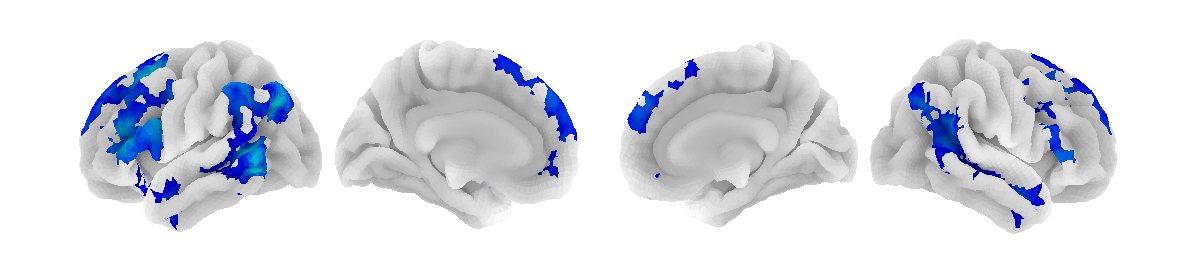}}
\textsf{B2. GPT-2.2 - LSTM.2}
\centerline{\includegraphics[width=.5\columnwidth]{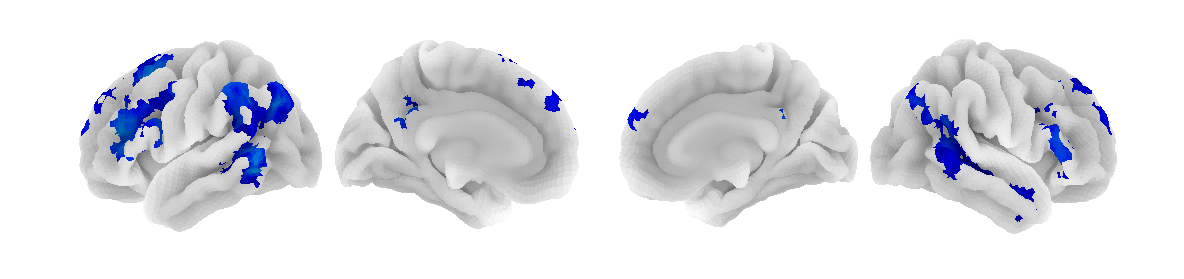}}
\textsf{B3. GPT-2.4 - LSTM.4}
\centerline{\includegraphics[width=.5\columnwidth]{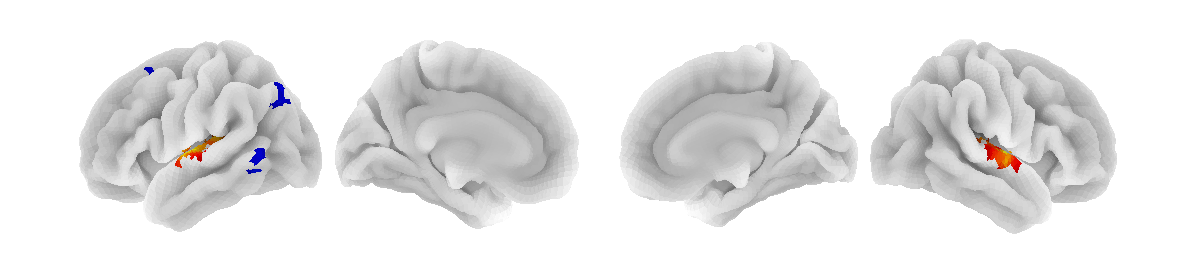}}
\vskip -0.05in
\textsf{C1. (GPT-2.1 $-$ LSTM.1) $\times$ (Trained $-$ Untrained)}
\centerline{\includegraphics[width=.5\columnwidth]{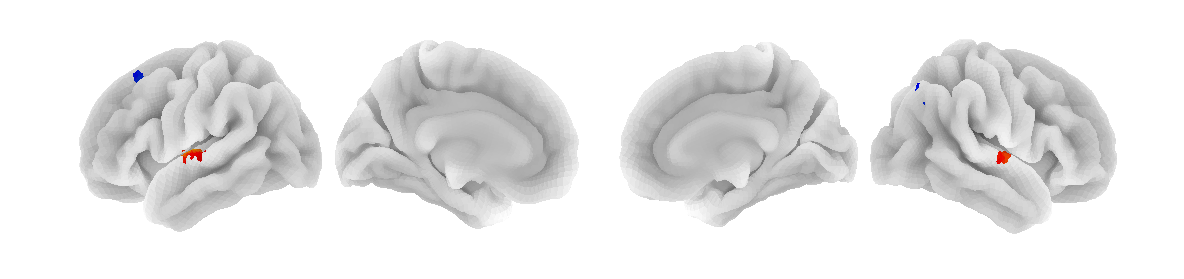}}
\textsf{C2. (GPT-2.2 $-$ LSTM.2) $\times$ (Trained $-$ Untrained)}
\centerline{\includegraphics[width=.5\columnwidth]{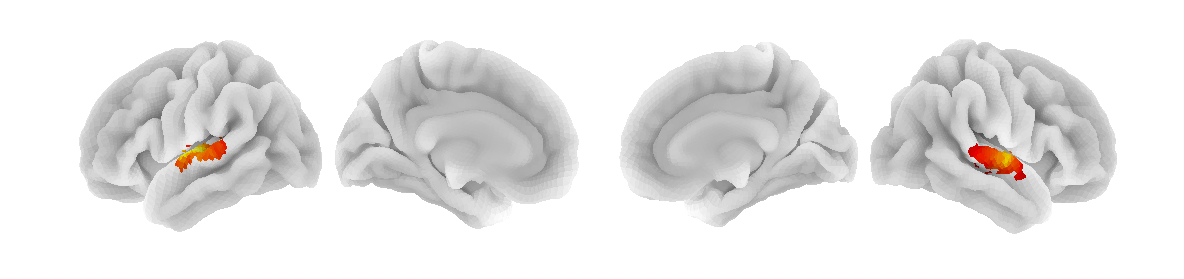}}
\textsf{C2. (GPT-2.4 $-$ LSTM.4) $\times$ (Trained $-$ Untrained)}
\centerline{\includegraphics[width=.5\columnwidth]{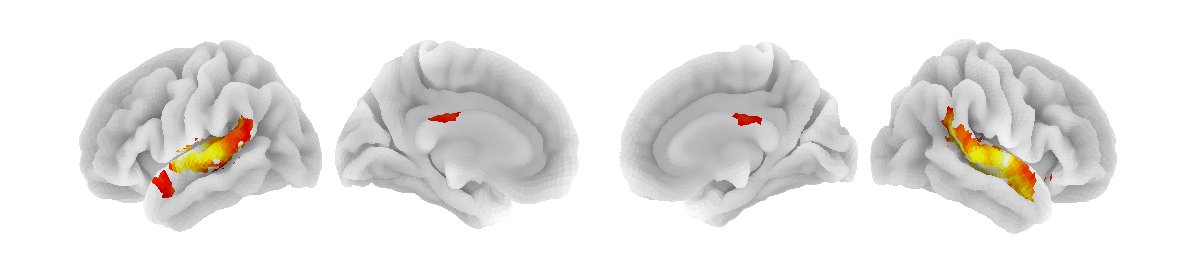}}
\centerline{\includegraphics[width=.4\columnwidth]{Figures/colorbars/colorbar_006.png}}
\caption{
\textbf{Panel A) LSTM vs. GloVe when trained on the small Wikipedia dataset.} Contrast maps for LSTM vs. GloVe. Results strongly depend on the training dataset (compare panel S6A and Figure 5A), with less brain regions identified with the small dataset. 
\textbf{Panel B) GPT-2 vs. LSTM when trained on the full dataset.} Contrasts maps for GPT-2 vs. LSTM for 1-layer, 2-layers and 4-layers models (respectively panels B1, B2 and B3). GPT-2 better predicts brain activity around the Heschel gyri, for the 4-layer version, but LSTM outperforms GPT-2 in most of the language network.\\
\textbf{Panel C) Interaction between model architecture (GPT-2 vs. LSTM) and Training (trained vs. untrained)} for 1-layer (top), 2-layers (middle) and 4-layer (bottom) models. GPT-2 benefits more from training than LSTM. The more layers, the more it learns.
}
\label{appendix:Figure:comparing_models}
\end{center}
\vskip -0.2in
\end{figure}

\clearpage

\clearpage
\onecolumn

\section{Relation Between Brain Score and Perplexity}

\setcounter{table}{2}
\begin{table}[h]
\caption{\textbf{Test Perplexities of models and their brains score in the SRM25 network.} For each model type, the best score is highlighted in bold. }

\vskip 0.15in
\begin{center}
\begin{small}
\begin{sc}
\begin{tabular}{lcrrr}
\toprule
Model & Training Dataset &  Dataset size & Perplexity & Brain Score \\
\midrule
LSTM L-1 H-768   & Wikipedia                        & 425M & 167.25    &   0.0828  \\
LSTM L-1 H-768   & Gut. small                       & 240M & 46.04     &   0.0973  \\
LSTM L-1 H-768   & Gut. medium                      & 737M & 39.39     &   0.0957  \\
LSTM L-1 H-768   & Gut. large                       & 2.2G & 35.76     &   0.0934  \\
LSTM L-1 H-768   & Gut. xlarge                      & 4.4G & 34.36     &   0.0920  \\
LSTM L-1 H-768   & Gut. xlarge + Wikipedia (Full)   & 4.8G & 35.40     &   0.0962  \\
GPT-2 L-1 H-768  & Gut. xlarge + Wikipedia (Full)   & 4.8G & 30.62     &   0.0795  \\
BERT L-1 H-768   & Gut. xlarge + Wikipedia (Full)   & 4.8G & 17.83     &   0.0898  \\
LSTM L-2 H-768   & Wikipedia                        & 425M & 160.03    &   0.0850  \\
LSTM L-2 H-768   & Gut. small                       & 240M & 42.67     &   0.0966  \\
LSTM L-2 H-768   & Gut. medium                      & 737M & 36.22     &   0.0974  \\
LSTM L-2 H-768   & Gut. large                       & 2.2G & 32.61     &   0.1032  \\
LSTM L-2 H-768   & Gut. xlarge                      & 4.4G & 31.21     &   0.0984  \\
LSTM L-2 H-768   & Gut. xlarge + Wikipedia (Full)   & 4.8G & 32.14     &   0.0981  \\
GPT-2 L-2 H-768  & Gut. xlarge + Wikipedia (Full)   & 4.8G & 26.22     &   0.0864  \\
BERT L-2 H-768   & Gut. xlarge + Wikipedia (Full)   & 4.8G & 11.57     &   0.0954  \\
LSTM L-4 H-768   & Gut. xlarge + Wikipedia (Full)   & 4.8G & 31.58     &   0.1002  \\
GPT-2 L-4 H-768  & Gut. xlarge + Wikipedia (Full)   & 4.8G & 23.62     &   0.0986  \\
\textbf{BERT L-4 H-768}   & \textbf{Gut. xlarge + Wikipedia (Full)}   & \textbf{4.8G} & \textbf{9.00}     &   \textbf{0.1057}  \\
\midrule
       ---       & Validation set                   & 1.1G & ---       &   ---      \\
\midrule
       ---       & Test set                         & 1.1G & ---       &   ---      \\

\bottomrule

\end{tabular}
\end{sc}
\end{small}
\end{center}
\end{table}

%
%

\end{document}